\documentclass[letterpaper]{article}
\usepackage{Advent}
\usepackage{times}
\usepackage{helvet}
\usepackage{courier}
\usepackage[hyphens]{url}
\usepackage{graphicx}
\usepackage{natbib}
\usepackage{caption}
\usepackage{algorithm}
\usepackage{algorithmic}
\usepackage{amsmath}
\usepackage{cleveref}
\usepackage{tabularx}
\usepackage[caption=true,font=normalsize,labelfont=sf,textfont=sf]{subfig}
\usepackage{multirow}
\usepackage{amsfonts}
\usepackage{float}
\usepackage{afterpage}
\usepackage{pifont}

\usepackage{newfloat}
\usepackage{listings}
\DeclareCaptionStyle{ruled}{labelfont=normalfont,labelsep=colon,strut=off}
\floatstyle{ruled}
\newfloat{listing}{tb}{lst}{}
\floatname{listing}{Listing}

\Crefname{figure}{Fig.}{Figs.}
\Crefname{equation}{Eq.}{Eqs.}
\Crefname{section}{Sec.}{Secs.}
\Crefname{subsection}{Sec.}{Secs.}
\crefname{subsection}{Sec.}{Secs.}
\crefname{section}{Sec.}{Secs.}

\def\eg{\emph{e.g.}}

\def\ie{\emph{i.e.}}

\newcommand{\xmark}{\ding{55}}

\title{Adverse Weather-Independent Framework Towards Autonomous Driving Perception through Temporal Correlation and Unfolded Regularization}
\author{
    Wei-Bin Kou\textsuperscript{\rm 1, \rm 2},
    Guangxu Zhu\textsuperscript{\rm 3},
    Rongguang Ye\textsuperscript{\rm 2},
    Jingreng Lei\textsuperscript{\rm 1},
    Shuai Wang\textsuperscript{\rm 4}, \\
    Qingfeng Lin\textsuperscript{\rm 1},
    Ming Tang\textsuperscript{\rm 2},
    Yik-Chung Wu\textsuperscript{\rm 1}
}
\affiliations{
    \textsuperscript{\rm 1}Department of Electrical and Electronic Engineering, The University of Hong Kong, Hong Kong, China.\\
    \textsuperscript{\rm 2}Department of Computer Science and Engineering, Southern University of Science and Technology, Shenzhen, China.\\
    \textsuperscript{\rm 3}Shenzhen International Center For Industrial And Applied Mathematics, Shenzhen Research Institute of Big Data, \\Shenzhen, China.\\
    \textsuperscript{\rm 4}Shenzhen Institutes of Advanced Technology, Chinese Academy of Sciences, Shenzhen, China.\\
}

\begin{document}

\maketitle

\begin{abstract}
Various adverse weather conditions such as fog and rain pose a significant challenge to autonomous driving (AD) perception tasks like semantic segmentation, object detection, etc. The common domain adaption strategy is to minimize the disparity between images captured in clear and adverse weather conditions. However, domain adaption faces two challenges: (I) it typically relies on utilizing clear image as a reference, which is challenging to obtain in practice; (II) it generally targets single adverse weather condition and performs poorly when confronting the mixture of multiple adverse weather conditions. To address these issues, we introduce a reference-free and \underline{Adv}erse weather condition-independ\underline{ent} (Advent) framework (rather than a specific model architecture) that can be implemented by various backbones and heads. This is achieved by leveraging the homogeneity over short durations, getting rid of clear reference and being generalizable to arbitrary weather condition. Specifically, Advent includes three integral components: (I) Locally Sequential Mechanism (LSM) leverages temporal correlations between adjacent frames to achieve the weather-condition-agnostic effect thanks to the homogeneity behind arbitrary weather condition; (II) Globally Shuffled Mechanism (GSM) is proposed to shuffle segments processed by LSM from different positions of input sequence to prevent the overfitting to LSM-induced temporal patterns; (III) Unfolded Regularizers (URs) are the deep unfolding implementation of two proposed regularizers to penalize the model complexity to enhance across-weather generalization. We take the semantic segmentation task as an example to assess the proposed Advent framework. Extensive experiments demonstrate that the proposed Advent outperforms existing state-of-the-art baselines with large margins.
\end{abstract}

\section{Introduction} \label{introduction}
Autonomous driving (AD) perception is a critical component of safe and reliable AD systems \cite{ma2024slowtrack,han2025dme,chen2024m,najibi2023unsupervised}, as it enables vehicles to understand and navigate their surroundings effectively. However, AD perception remains a significant challenge in adverse weather conditions like fog, rain, snow, etc., where visibility is reduced and sensor data becomes unreliable \cite{li2023domain,alzanin2025explainable,hao2024your}. These conditions can severely degrade the performance of AD perception models \cite{wang2024deepaccident,li2024light} like object detection model, semantic segmentation model, etc. For instance, dense fog obscures road boundaries and heavy rain introduces motion blur \cite{bi2024learning,huang2024sunshine}, leading to a significant drop in prediction accuracy for AD perception systems. In this paper, we particularly focus on semantic segmentation, a fundamental task in AD perception that assigns a class label to every pixel in an image, as it is crucial to understand the driving environment.

Currently, significant progresses have been made in improving AD semantic segmentation task under adverse weather conditions \cite{mirza2022efficient,kalb2023principles}. A common strategy is domain adaptation \cite{zhao2024unimix,fu2024auxiliary}, which reduces the domain gap between images captured in clear weather condition (serving as references) and various adverse weather conditions. However, domain adaption faces two challenges: (I) obtaining those reference images in real-world driving scenarios is often impractical due to the dynamic nature of weather and the unavailability of paired data. (II) existing domain adaption methods are typically designed to address specific weather conditions (\eg, fog), and fail to generalize well to other adverse conditions (\eg, rain or snow) that are not fine-tuned. 

\begin{figure*}[tp]
\centering 
\includegraphics[width=\linewidth]{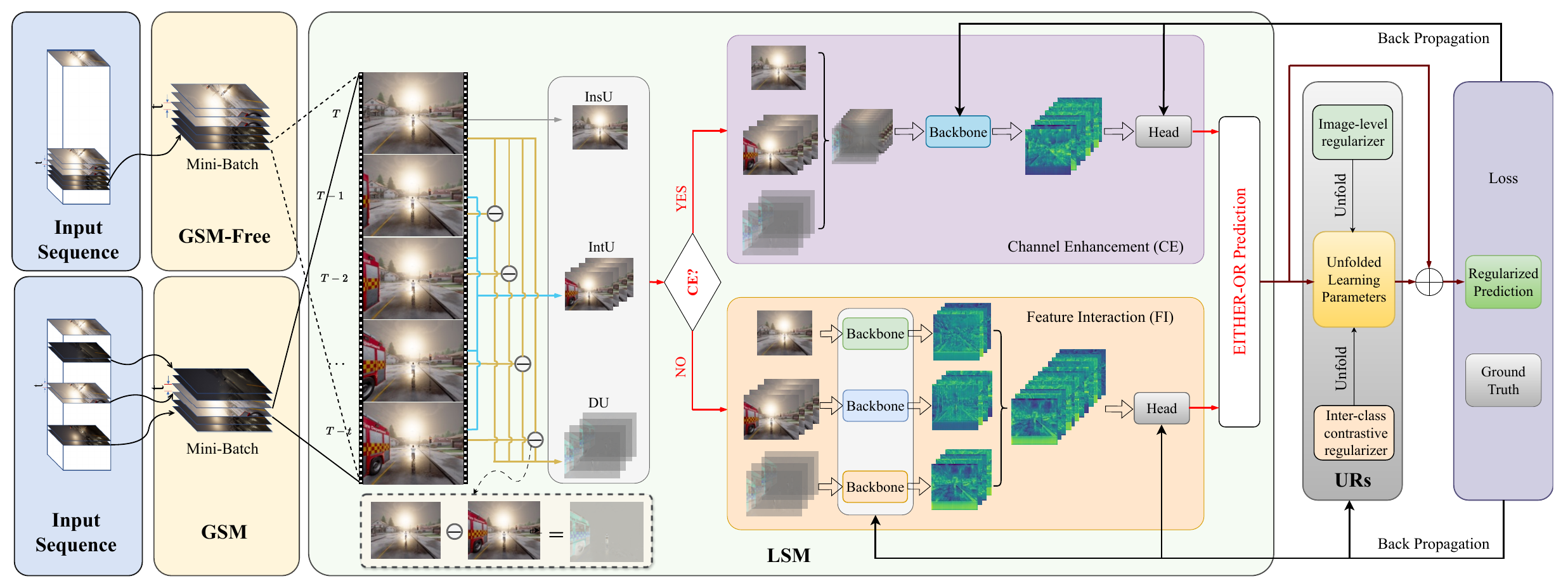}
\caption{Overview of the proposed Advent framework. LSM captures the temporal correlations by integrating InsU, IntU, and DU to sense immediate scenes, stable background, and dynamic changes, respectively. GSM shuffles segments processed by LSM from different positions of input sequence to avoid the overfitting to LSM-induced temporal patterns, whereas GSM-Free does not. URs convert heuristic-searched regularizers' coefficients in optimization loss to learnable parameters to avoid suboptimal performance. LSM, GSM, and URs work together to achieve the adverse weather condition-agnostic effect. Notably, involved backbone(s), head, and URs' tunable parameters are optimized by back propagation unifirmly.}
\label{Fig.LSM_with_EF}
\end{figure*}

To address these challenges, we propose a reference-free and \underline{Adv}erse weather condition-independ\underline{ent} (Advent) framework. Unlike previous approaches that rely on correlations of adverse weather conditions against clear weather condition, the intuition behind Advent is to exploit the temporal correlations presented in consecutive frames of driving environments, as frames of arbitrary weather tend to exhibit homogeneity and inter-dependence over short durations. By leveraging these temporal patterns, our approach does not rely on clear reference and is generalizable to arbitrary adverse weather condition. This is achieved by following integral components: (I) Locally Sequential Mechanism (LSM): LSM captures temporal correlations from consecutive frames by utilizing three distinct units: Instant Unit (InsU) is proposed to detect instantaneous scenes of driving environments, such as traffic light, pedestrians, and vehicles; Integral Unit (IntU) is introduced to extract stable and shared background information from the scene, such as weather conditions and light conditions; Derivative Unit (DU) is presented to focus on dynamic changes, such as moving objects or shifting weather patterns. Based on the cooperation of these three units, LSM can effectively understand the driving environment regardless of what kind of the weather it is. (II) Globally Shuffled Mechanism (GSM): GSM is presented to mitigate the overfitting caused by specific input sequence orders in LSM. This is implemented by rearranging segments processed by LSM from different positions from input sequence, ensuring the model learns generalizable patterns across varied weather sequences. (III) Unfolded Regularizers (URs): URs are proposed to enhance generalization across diverse adverse weather conditions by penalizing model complexity. Specifically, we first propose an image-level regularizer and an inter-class contrastive regularizer. We then use the deep unfolding technique \cite{monga2021algorithm} to unfold these two regularizers, through which two regularizer weights in optimization objective are converted into learnable parameters to avoid instability and suboptimal performance caused by heuristics or exhaustive searches. These learnable parameters are finally tuned along with the model parameters together. The proposed Advent framework is illustrated in \Cref{Fig.LSM_with_EF}. We take the semantic segmentation task as an example to assess the proposed Advent framework. Extensive experiments show that Advent achieves strong generalization across a variety of adverse weather conditions, and it surpasses baselines significantly.

The main contributions of this work are highlighted as follows:
\begin{itemize}
    \item The proposed Advent considers both local temporal correlations by LSM to get rid of the dependence on clear reference and to generalize well to arbitrary adverse weather condition, and global randomness by GSM to prevent overfitting to LSM-induced temporal patterns.
    \item Advent also presents an image-level regularizer and an inter-class contrastive regularizer to enhance the model's across-weather generalization, and additionally unfolds both regularizers into layers by deep unfolding technique. This can convert weights of both regularizers in optimization objective to learnable parameters instead of heuristic-search hyperparameters to avoid instability and suboptimal performance.
    \item We take semantic segmentation task as example to assess Advent. Extensive experiments demonstrate that Advent can generalize well under various adverse weather conditions, and outperforms baselines significantly. 
\end{itemize}

\section{Related Work}

\subsection{Domain Adaption for AD Perception}
AD perception under various adverse conditions has recently attracted increasing attention due to the strict safety demand in various outdoor perception tasks \cite{rothmeier2024time,zhang2023perception,yang2025uawtrack}. Domain adaption is the major scheme in enhancing AD perception capabilities under various adverse conditions \cite{hoyer2023domain,yang2024semantic,li2023vblc,jiang2024domain}. Some studies in domain adaption aim to learn a generalizable model from single or multiple related but distinct source domains where target data is inaccessible during model learning \cite{zhou2022domain,Rothmeier_2024_WACV}. Some adversarial training-based domain adaption works attempt to narrow the domain gap via style transfer or learning indistinguishable representations \cite{kim2020learning,huang2024sunshine}. Furthermore, equipped with abundant data from various domains, curriculum learning-based domain adaption realize a progressive adaptation towards a distant target domain \cite{karim2023c,wang2024curriculum}. In addition, extending domain adaptation to multiple target domains usually employs domain transfer \cite{yang2024generalized,li2024di} to enhance generalization across domains. While domain adaption achieves remarkable advancements for AD perception under adverse weather conditions, it generally faces intricate challenges, such as relying on clear reference and performing poorly when confronting the mixture of multiple adverse weather conditions. In this work, unlike domain adaption, the proposed Advent can sense instantaneous scenes, stable background, and dynamic changes by considering temporal correlations, which is reference-free and robust to arbitrary adverse conditions.

\subsection{Street Scene Semantic Segmentation}
Semantic segmentation is dedicated to equipping machines with the ability to interpret environments surrounding vehicles \cite{kim2024weakly,tang2025unleashing,gao2024generalize}. This understanding typically derives from diverse sensory inputs, including images and lidars \cite{kim2024weakly,lin2024bev}. Such capabilities are indispensable for AD as they facilitate comprehension of street layouts, encompassing roads, pedestrians, sidewalks, buildings, and various other static and dynamic components. The evolution of semantic understanding has been profoundly influenced by advancements in machine learning (ML), especially through the deployment of deep learning (DL) methods. Initially, the adoption of Fully Convolutional Networks (FCNs)-based models marked a significant enhancement in performance \cite{yang2022deaot,zhou2022rethinking,chen2018encoderdecoder,badrinarayanan2017segnet,yu2021bisenet,xiao2023baseg}. More recently, Transformer-based methods \cite{xie2021segformer,zhang2022topformer,wan2023seaformer,song2021attanet} have been introduced for semantic segmentation tasks. In this work, we enhance AD semantic segmentation based on DL backbones and heads by considering temporal correlation and unfolded regularization techniques.

\begin{figure}[tp]
\centering 
\includegraphics[width=\linewidth]{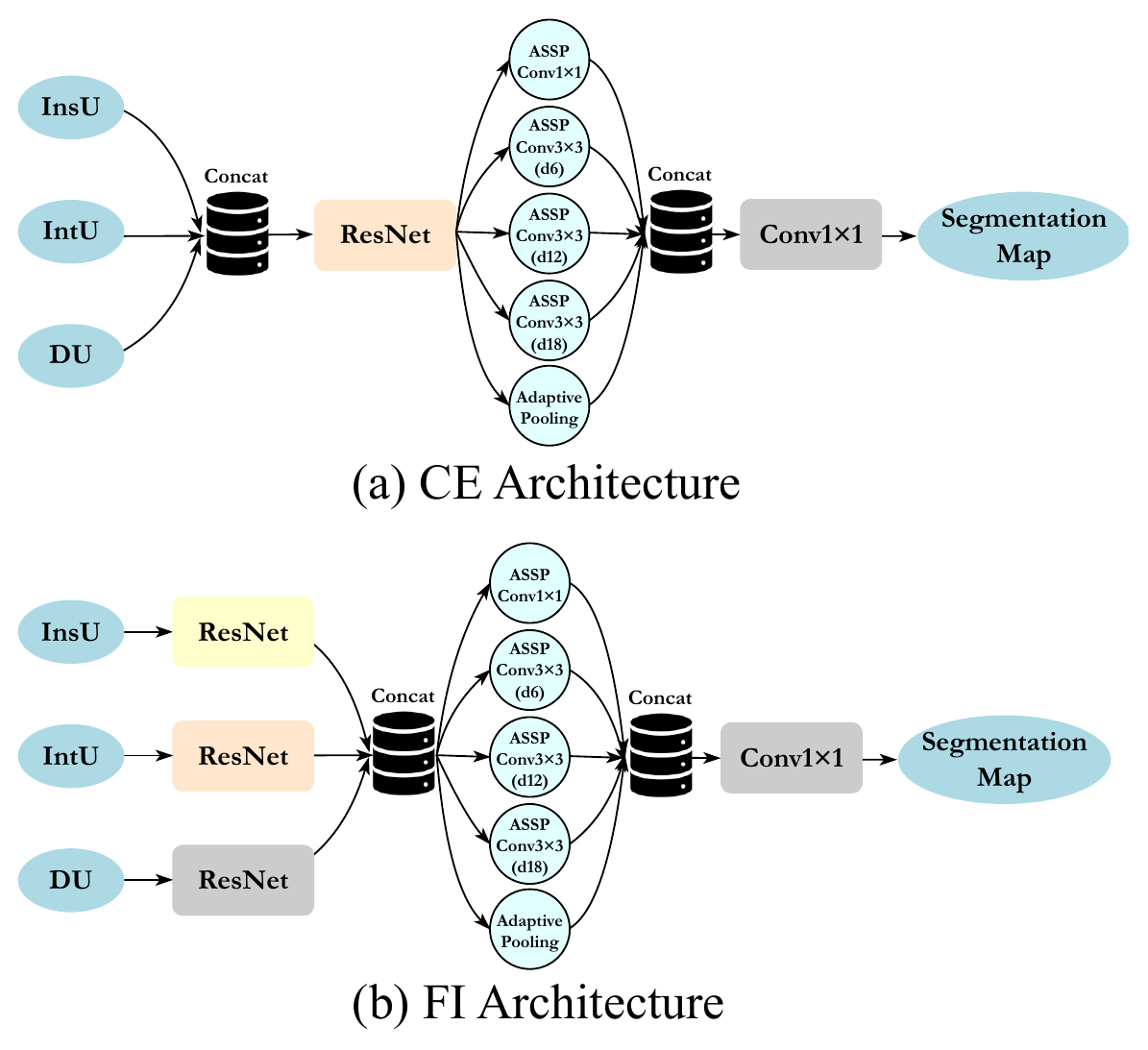}
\caption{Architecture Comparison of CE and FI, taking ResNet \cite{He_2016_CVPR} as backbone and ASSP \cite{chen2018encoderdecoder} as downstream head.}
\label{Fig.EF_vs_LF}
\end{figure}

\section{Methodology}
We firstly introduce Locally Sequential Mechanism (LSM). Subsequently, we present Globally Shuffled Mechanism (GSM). Finally, we detail Unfolded Regularizers (URs). 

\subsection{Locally Sequential Mechanism (LSM)} \label{sec:LSaGS}
LSM captures temporal correlations across consecutive frames using three specialized units: the Instant Unit (InsU), the Integral Unit (IntU), and  the Derivative Unit (DU). Specifically, InsU is designed to process the current image in captured sequence as vehicle drives. It extracts instantaneous scenes, such as traffic lights, pedestrians, and vehicles. This is crucial for the vehicle to accurately understand immediate surroundings. IntU focuses on processing stable background information from the environment like weather conditions and lighting conditions, achieved by analyzing past consecutive images prior to the current frame. This enables IntU to understand what remains constant over time, which is important for consistent understanding of the scene. DU analyzes dynamic changes (such as moving objects or shifting weather patterns) in the environment by comparing the current frame against previous ones. By integrating these three units together, LSM achieves a comprehensive understanding of the driving environment, regardless of what the weather condition it is. Notably, LSM depth is used to indicate the count of past consecutive frames that are considered. 

\begin{figure*}[t]
\centering
\includegraphics[width=0.85\linewidth,height=0.204\linewidth]{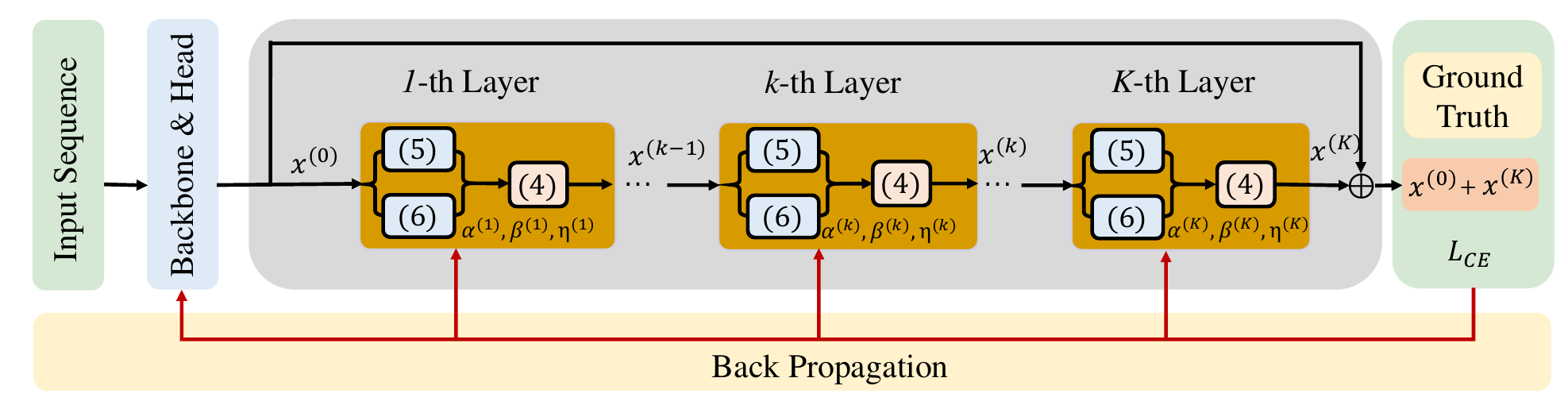}
\caption{Illustration of the unfolding and optimization process of URs. (4), (5), and (6) represent equation number.}
\label{Fig.RDU}
\end{figure*}

Channel Enhancement (CE) and Feature Interaction (FI) are two policies for fusing features from InsU, IntU, and DU. In CE strategy, features from InsU, IntU, and DU are first merged, then these merged features are processed by a shared backbone model. This integration allows the backbone model to fully utilize the correlations between these different types of features, enhancing the model for understanding scenes. The shared backbone-processed features are then passed to downstream head to generate prediction. On the other hand, in FI strategy, InsU, IntU, and DU process the their own features by three separate backbone models. This allows each type of features to be finely analyzed on its own. After processing, these different types of features are interacted to enhance downstream head's prediction accuracy. The philosophy behind CE and FI is illustrated in \Cref{Fig.LSM_with_EF}, and example model architectures of CE and FI are illustrated in \Cref{Fig.EF_vs_LF}. Ultimately, the either-or aggregation of above both predictions serves as the final prediction. 

\subsection{Globally Shuffled Mechanism (GSM)} \label{sec:GSM}
GSM is introduced to shuffle segments processed by LSM from different positions of input sequence into mini-batches. By doing so, each mini-batch reflects the diversity of the input sequence instead of following a strict sequence order. This avoids the overfitting to specific LSM-induced temporal patterns and enables the model to learn more generalizable and robust patterns, enhancing its prediction accuracy and generalization. Based on the shuffled mini-batches by GSM from the input sequence, the involved backbones and the downstream perception head are then optimized by back propagation. GSM is illustrated in the left part of \Cref{Fig.LSM_with_EF}.

In summary, we algorithmize the combination of GSM and LSM in \textit{Appendix I of Supplementary Materials}.

\begin{algorithm}[t]
\caption{Unfolding and Optimization of URs}
\label{Algo:DUN}
\begin{algorithmic}[1]
\REQUIRE $X$ (Input Sequence), $Y$ (Ground Truth), $K$ (Layer Number)
\ENSURE Model $\langle Backbone, Head \rangle$, Learnable variables $\{\alpha^{(k)}, \gamma^{(k)}, \eta^{(k)}\}_{k=1}^K$
\STATE Initialize model $\langle Backbone, Head \rangle \gets \langle B_0, H_0 \rangle$, and  $\{\alpha^{(k)}, \gamma^{(k)}, \eta^{(k)}\}_{k=1}^K \gets \alpha_0, \gamma_0, \eta_0$
\FOR{$epoch\ i \gets 1$ to $max\_epochs$}
    \STATE $x^{(0)} \gets \langle Backbone, Head \rangle(X)$
    \FOR{$layer\ k \gets 1$ to $K$}
        \STATE $\nabla_x L_{BD}(x^{(k-1)}) \gets \Cref{Eq:grad_L_BD}$
        \STATE $\nabla_x L_{\text{con}}(x^{(k-1)}) \gets \Cref{Eq:grad_L_con}$
        \STATE $x^{(k)} \gets \Cref{Eq:RDU_update}$
    \ENDFOR
    \STATE $x \gets x^{(0)} + x^{(K)}$
    \STATE $L \gets L_{CE}(x, Y)$
    \STATE $\langle Backbone, Head \rangle, \{\alpha^{(k)}, \gamma^{(k)}, \eta^{(k)}\}_{k=1}^K \gets L.Backward()$
\ENDFOR
\end{algorithmic}
\end{algorithm}

\subsection{Unfolded Regularizers (URs)} \label{sec:URs}
On top of the conventional pixel-level cross entropy loss (denoted as $L_{CE}$) in semantic segmentation, we propose an image-level regularizer and an inter-class contrastive regularizer to enhance the model's across-weather generalization. URs are the deep unfolding implementation of both regularizers, which aims at avoiding the heuristics and exhaustive searches on weights of them in optimization loss.

\subsubsection{Definitions of the proposed Regularizers.}
We denote the distrituion of predicted masks $X$ and the ground truth $Y$ as $P_X$ and $P_Y$, respectively. We propose to use Bhattacharyya distance (BD) \cite{bhattacharyya1943measure} as the image-level regularizer to quantify the distribution similarity between $X$ and $Y$. The definition of the BD regularizer over pixel $x$ is
\begin{align}
L_{BD} = D_B(P_X, P_Y) = -\ln(\sum\nolimits_x \sqrt{P_X(x) P_Y(x)}).
\end{align}

On the other hand, we propose to use InfoNCE \cite{oord2018representation} as an inter-class contrastive regularizer that can pull pixles of same class closer while push pixles of different classes apart. InfoNCE regularizer is defined as
\begin{align}
\hspace{-0.2cm}L_{con}\!=\!-\!\log\!\frac{\sum_{m=1}^{|\mathcal{U}|}\exp(\frac{\sigma(x, u_{m})}{\tau})}{\sum_{m=1}^{|\mathcal{U}|}\!\exp(\frac{\sigma(x, u_{m})}{\tau})\!\!+\!\!\sum_{n=1}^{|\mathcal{V}|}\!\exp(\frac{\sigma(x, v_n)}{\tau})},
\end{align}
where \(x\) represents the iterable anchor pixel in the predicted masks, \(u_m\) is any pixel from the same-class set \(\mathcal{U}\), \(v_n\) is any pixel from the different-class set \(\mathcal{V}\), \(\sigma(\cdot, \cdot)\) denotes the inner product, and \(\tau\) controls the separation sharpness.

\begin{table*}[t]
\centering
\setlength{\tabcolsep}{3.0pt}
\begin{tabular}{cccccccccc}
\hline
\multirow{2}{*}{Methods} & \multicolumn{4}{c}{Apolloscape}                   &  & \multicolumn{4}{c}{CARLA\_ADV}                    \\ \cline{2-5} \cline{7-10} 
                         & mIoU       & mPre       & mRec       & mF1        &  & mIoU       & mPre       & mRec       & mF1        \\ \hline
DeepLabv3+  & \underline{26.58$\pm$0.49} & 30.23$\pm$0.89 & \underline{31.14$\pm$0.57} & \underline{32.32$\pm$0.59} &  & \underline{36.90$\pm$1.00} & 46.96$\pm$0.97 & 42.44$\pm$0.79 & 43.78$\pm$0.93 \\
BiSeNetV2  & 22.92$\pm$0.86 & 27.85$\pm$1.20 & 27.10$\pm$1.11 & 27.12$\pm$1.57 &  & 28.80$\pm$1.93 & 34.47$\pm$2.23 & 33.46$\pm$2.19 & 33.59$\pm$2.30 \\
SegNet  & 21.01$\pm$0.51 & 24.80$\pm$1.11 & 24.96$\pm$0.54 & 24.60$\pm$0.70 &  & 31.67$\pm$2.11 & 38.45$\pm$3.09 & 36.55$\pm$2.44 & 37.15$\pm$2.70 \\
AttaNet  & 20.59$\pm$0.18 & 25.88$\pm$0.31 & 25.38$\pm$0.22 & 25.55$\pm$0.19 &  & 27.12$\pm$1.40 & 33.41$\pm$1.68 & 32.90$\pm$1.65 & 32.60$\pm$1.57 \\
BASeg  & 20.14$\pm$0.18 & 31.72$\pm$0.29 & 25.26$\pm$0.20 & 25.75$\pm$0.31 &  & 36.36$\pm$0.76 & \underline{57.16$\pm$0.43} & \underline{43.86$\pm$0.82} & \underline{45.01$\pm$0.71} \\
HRDA  & 21.55$\pm$0.27 & \underline{34.13$\pm$0.31} & 26.09$\pm$0.34 & 26.40$\pm$0.31 &  & 33.66$\pm$0.14 & 50.87$\pm$1.04 & 39.23$\pm$1.30 & 41.57$\pm$1.23 \\
SeaFormer  & 20.34$\pm$0.16 & 25.33$\pm$0.21 & 25.03$\pm$0.13 & 24.26$\pm$0.16 &  & 27.85$\pm$0.64 & 35.80$\pm$0.55 & 32.58$\pm$0.58 & 33.48$\pm$0.56 \\
TopFormer  & 20.41$\pm$0.00 & 25.19$\pm$0.45 & 25.20$\pm$0.22 & 24.28$\pm$0.20 &  & 29.84$\pm$1.73 & 38.09$\pm$2.72 & 34.29$\pm$1.75 & 35.41$\pm$2.08 \\
\textbf{Advent (Ours)}           &  \textbf{59.35$\pm$0.57} & \textbf{81.38$\pm$2.22} & \textbf{60.76$\pm$0.49} & \textbf{65.28$\pm$0.55} &  & \textbf{69.58$\pm$1.79} & \textbf{85.71$\pm$2.01} & \textbf{70.78$\pm$1.56} & \textbf{75.01$\pm$1.56} \\ \hline
\end{tabular}
\caption{Quantitative performance comparison of Advent against other baselines.}
\label{Tab.perf_comp}
\end{table*}

\begin{table*}[!htbp]
\centering
\renewcommand{\arraystretch}{0.24}
\addtolength{\tabcolsep}{-0.45pt}
\begin{tabularx}{\linewidth}{|llllll|}
\hline
\hspace{0.6cm}Raw Images &\hspace{0.3cm}Ground Truth &\hspace{0.6cm}BASeg &\hspace{0.6cm}HRDA &\hspace{0.3cm}DeepLabv3+ &\hspace{0.1cm}\textbf{Advent (Ours)} \\
\hline
\includegraphics[width=0.162\linewidth, height=0.11\linewidth]{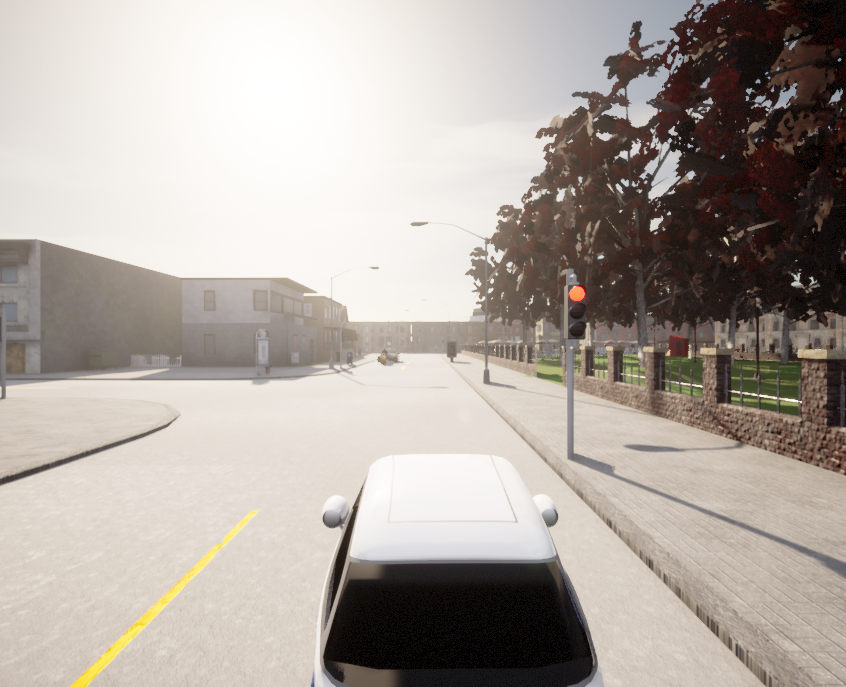} &\hspace{-0.47cm}
\includegraphics[width=0.162\linewidth, height=0.11\linewidth]{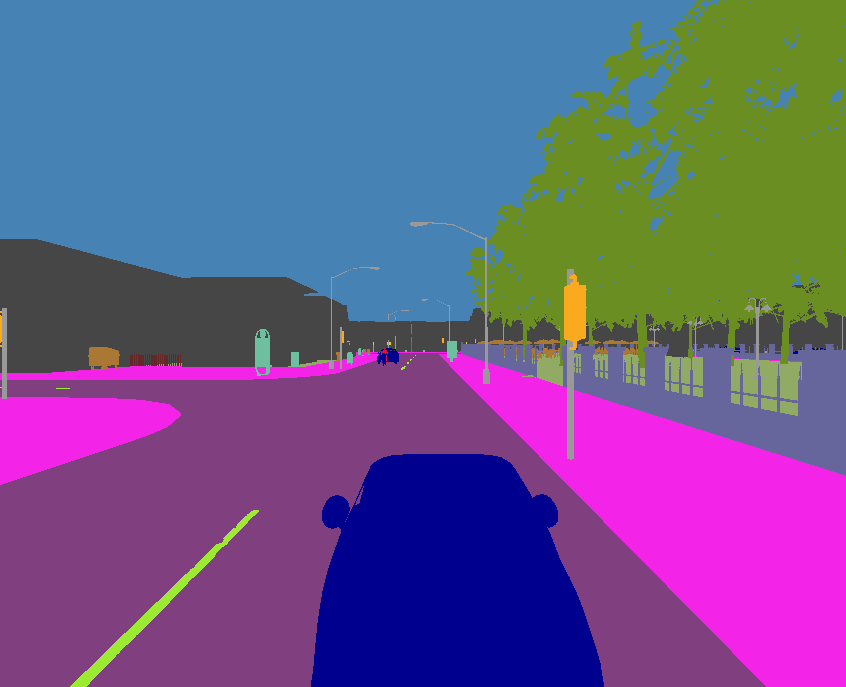} &\hspace{-0.47cm}
\includegraphics[width=0.162\linewidth, height=0.11\linewidth]{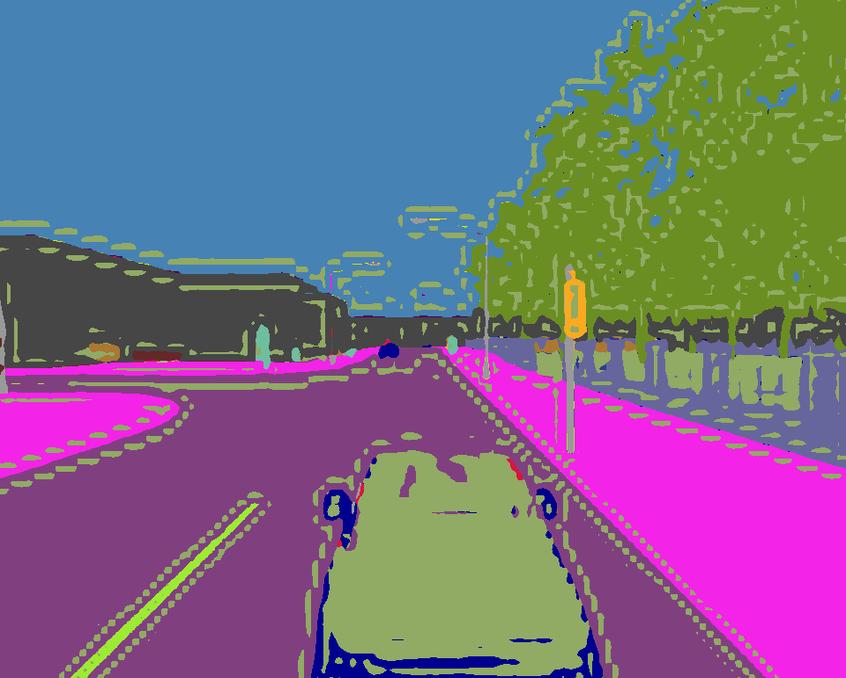} &\hspace{-0.47cm}
\includegraphics[width=0.162\linewidth, height=0.11\linewidth]{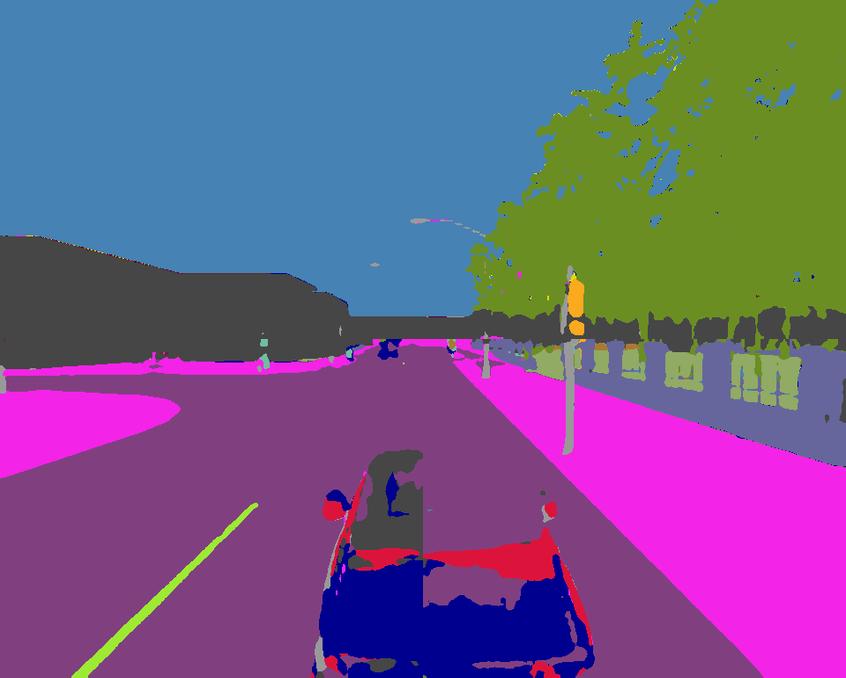} &\hspace{-0.47cm}
\includegraphics[width=0.162\linewidth, height=0.11\linewidth]{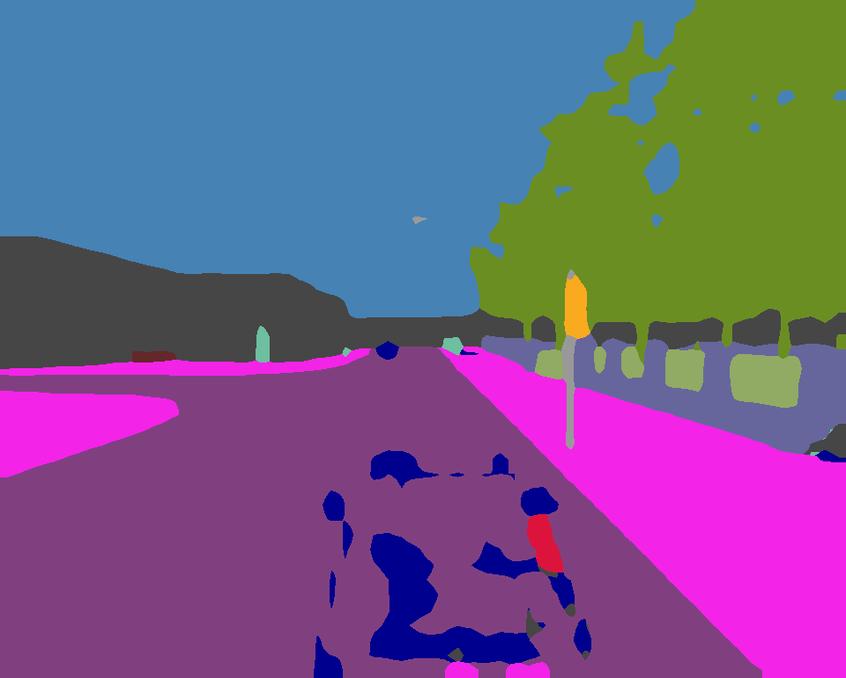} &\hspace{-0.43cm}
\includegraphics[width=0.162\linewidth, height=0.11\linewidth]{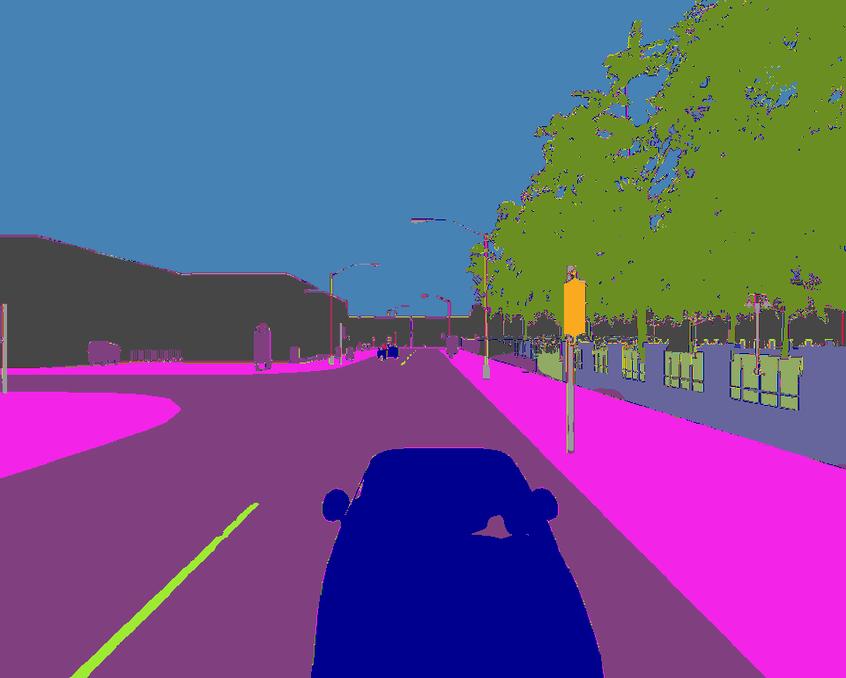}\\
\hline

\includegraphics[width=0.162\linewidth, height=0.11\linewidth]{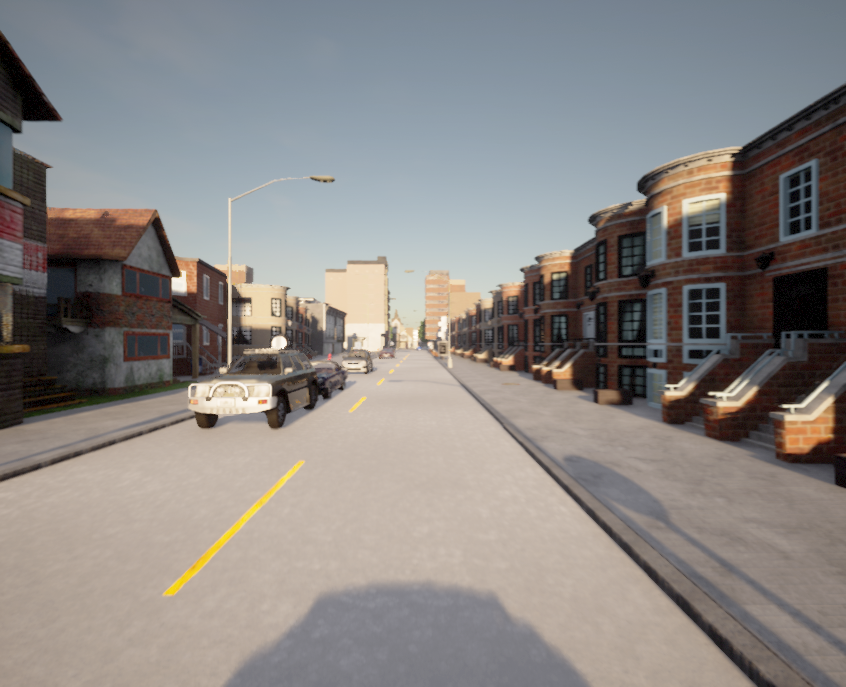} &\hspace{-0.47cm}
\includegraphics[width=0.162\linewidth, height=0.11\linewidth]{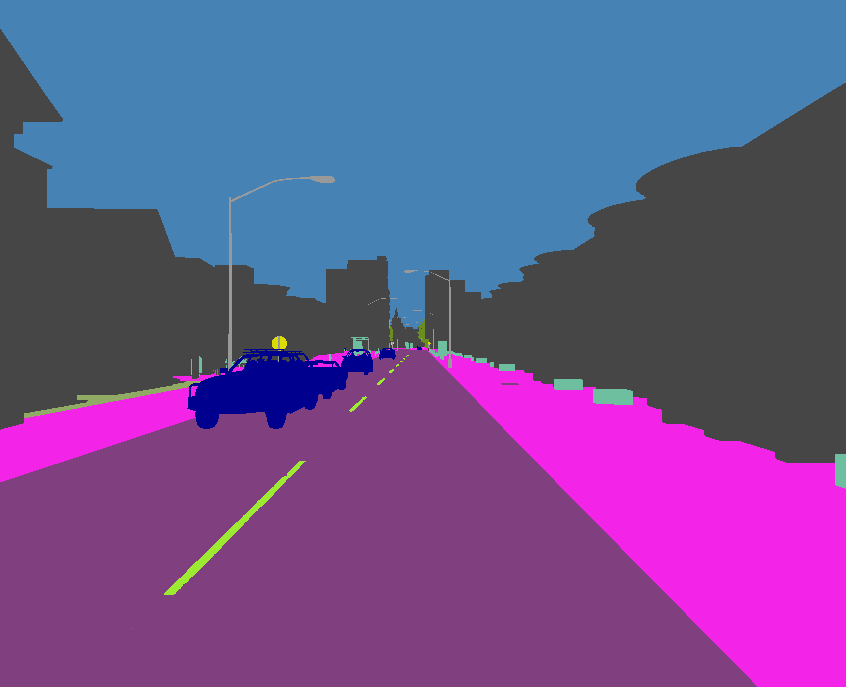} &\hspace{-0.47cm}
\includegraphics[width=0.162\linewidth, height=0.11\linewidth]{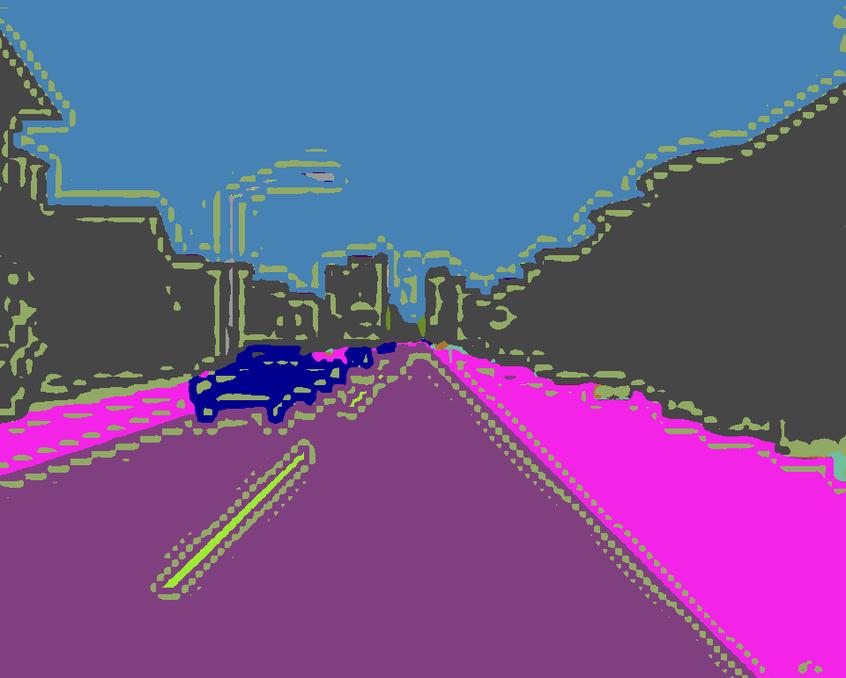} &\hspace{-0.47cm}
\includegraphics[width=0.162\linewidth, height=0.11\linewidth]{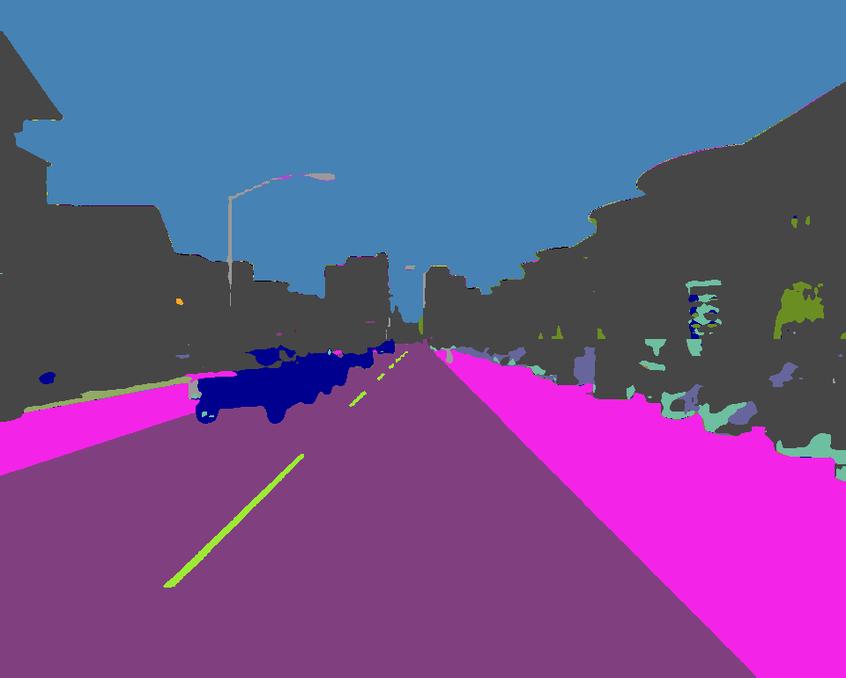} &\hspace{-0.47cm}
\includegraphics[width=0.162\linewidth, height=0.11\linewidth]{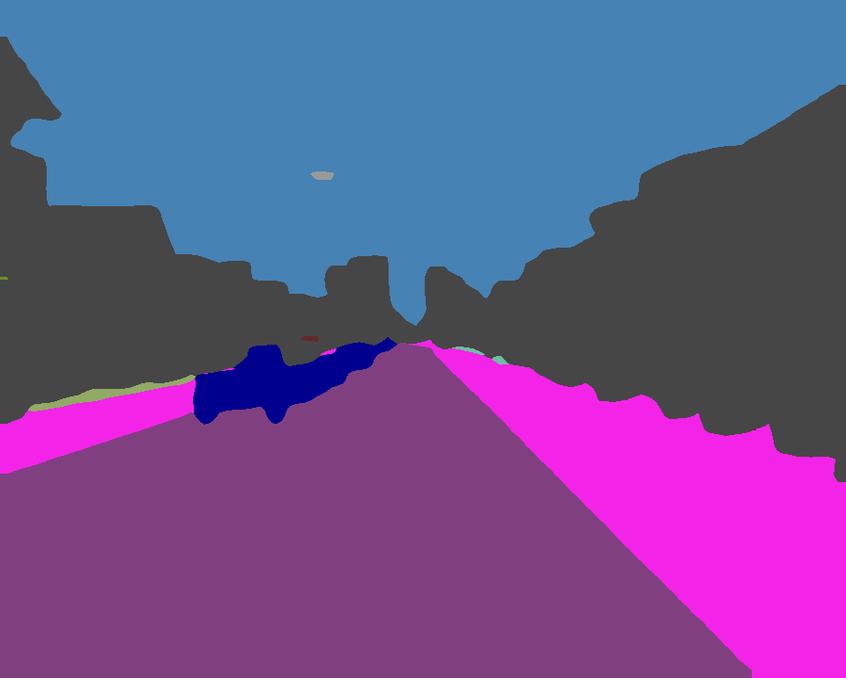} &\hspace{-0.43cm}
\includegraphics[width=0.162\linewidth, height=0.11\linewidth]{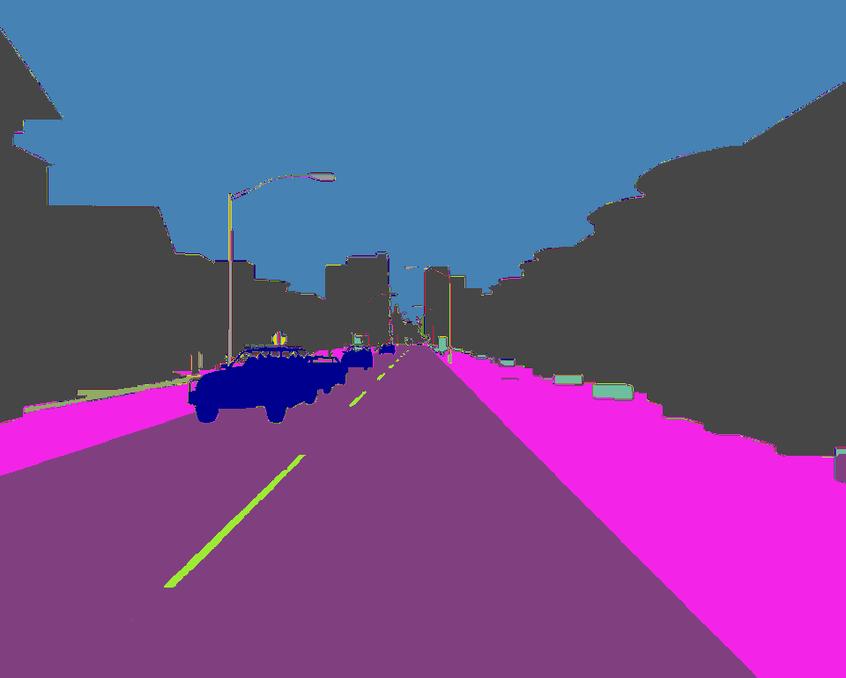}\\
\hline

\includegraphics[width=0.162\linewidth, height=0.11\linewidth]{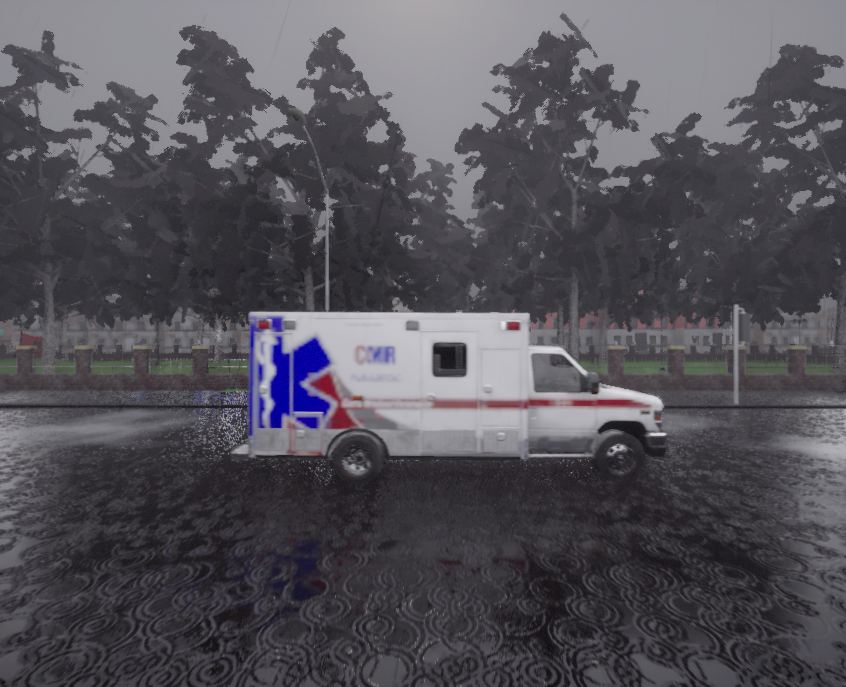} &\hspace{-0.47cm}
\includegraphics[width=0.162\linewidth, height=0.11\linewidth]{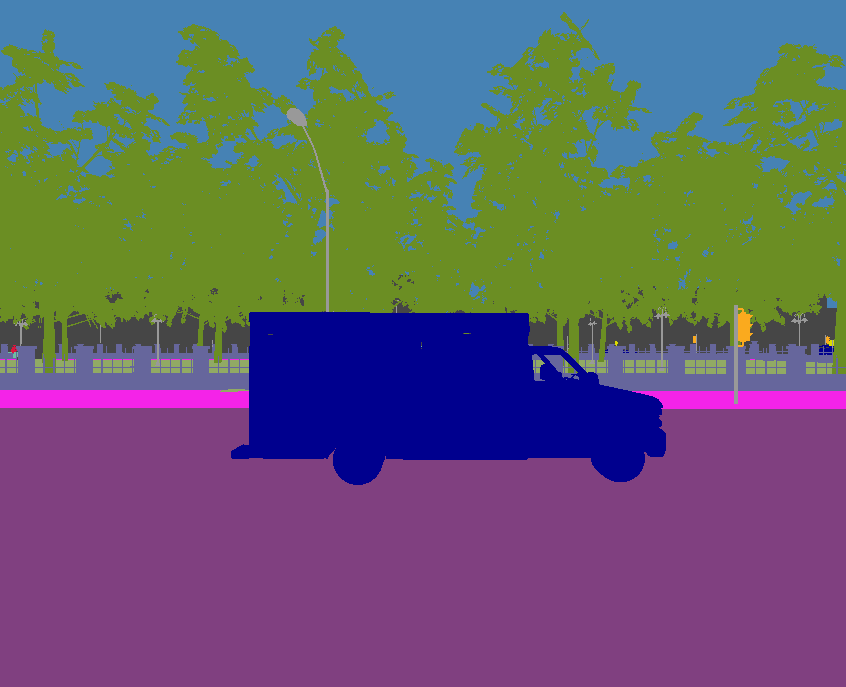} &\hspace{-0.47cm}
\includegraphics[width=0.162\linewidth, height=0.11\linewidth]{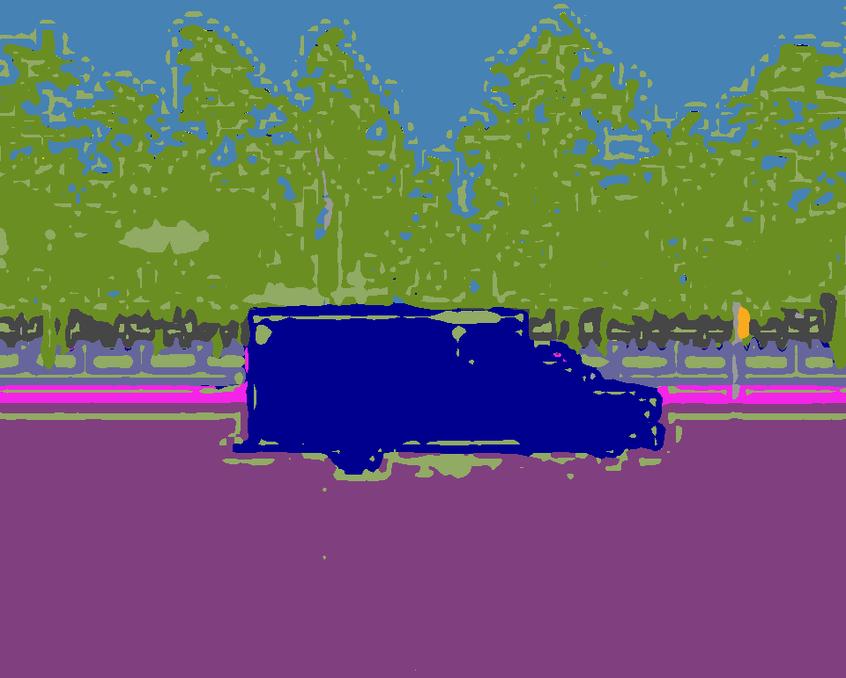} &\hspace{-0.47cm}
\includegraphics[width=0.162\linewidth, height=0.11\linewidth]{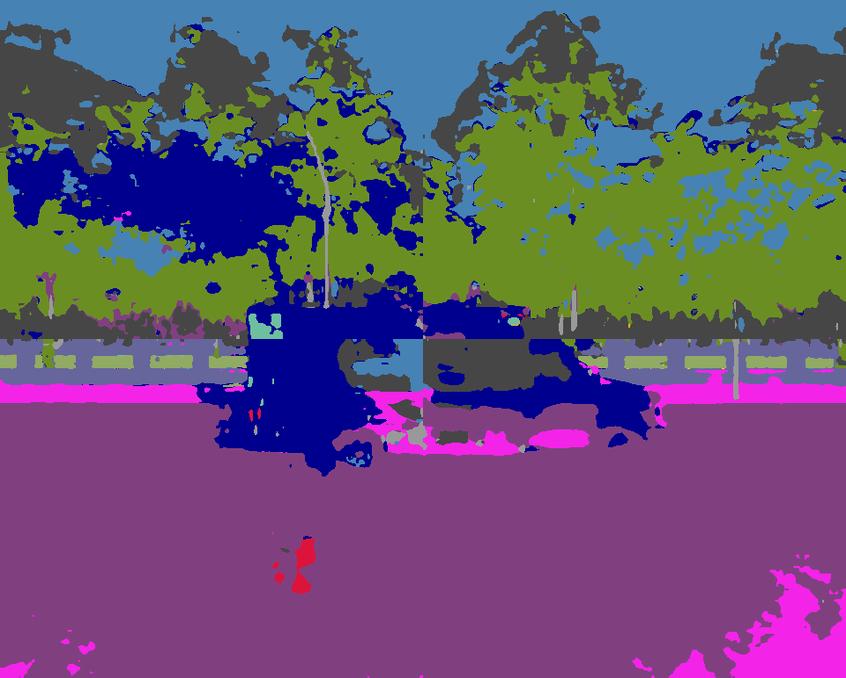} &\hspace{-0.47cm}
\includegraphics[width=0.162\linewidth, height=0.11\linewidth]{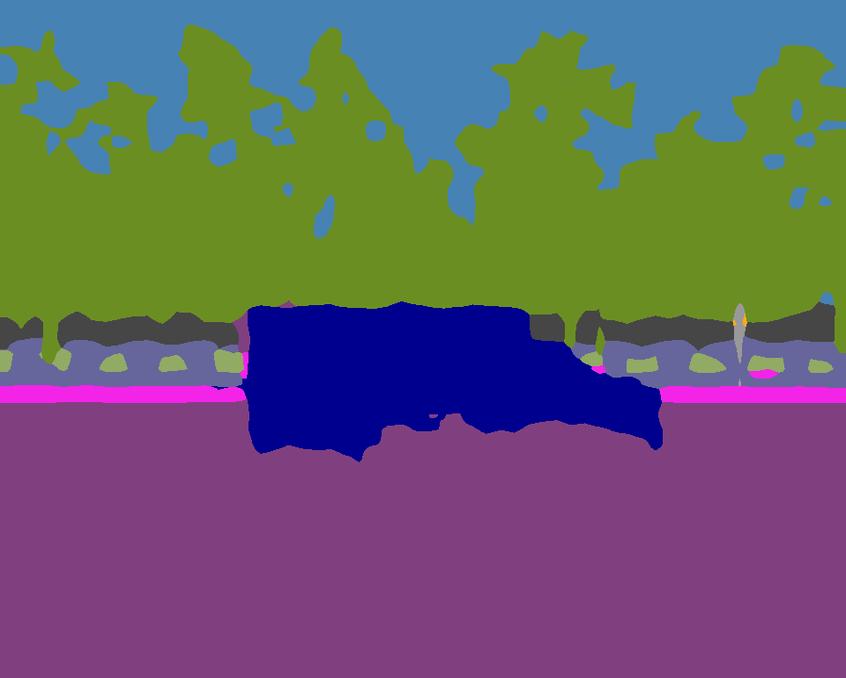} &\hspace{-0.43cm}
\includegraphics[width=0.162\linewidth, height=0.11\linewidth]{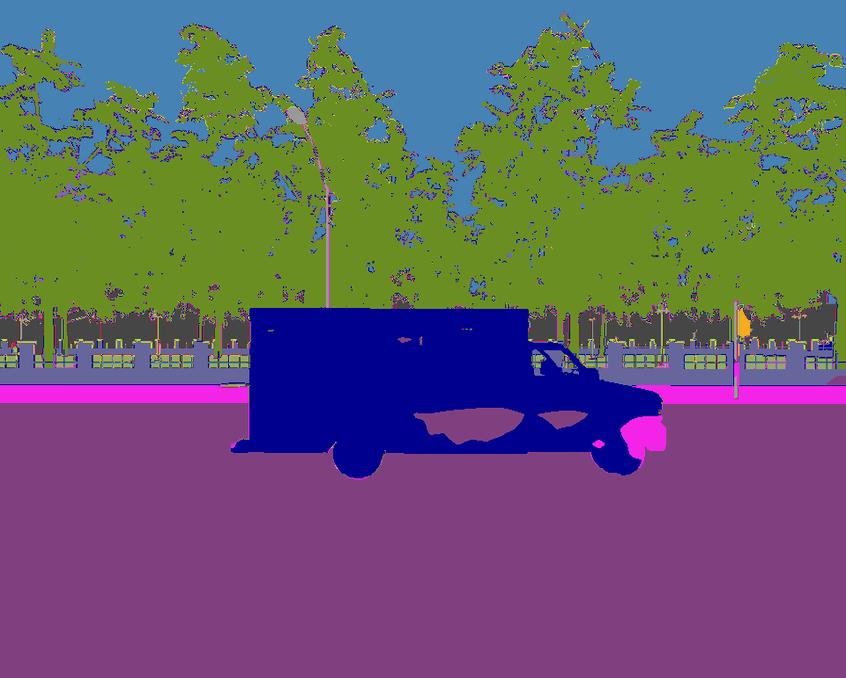}\\
\hline

\includegraphics[width=0.162\linewidth, height=0.11\linewidth]{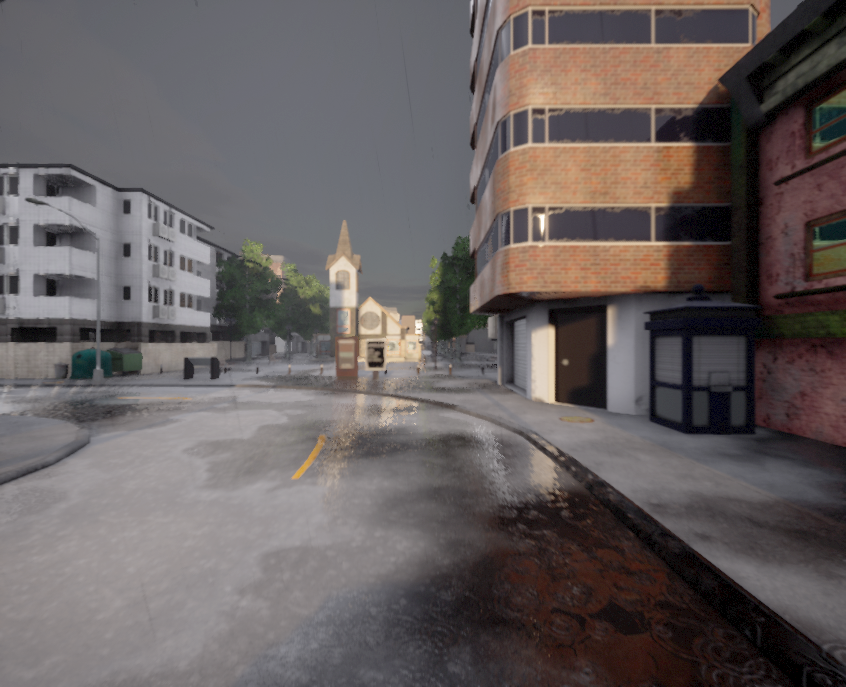} &\hspace{-0.47cm}
\includegraphics[width=0.162\linewidth, height=0.11\linewidth]{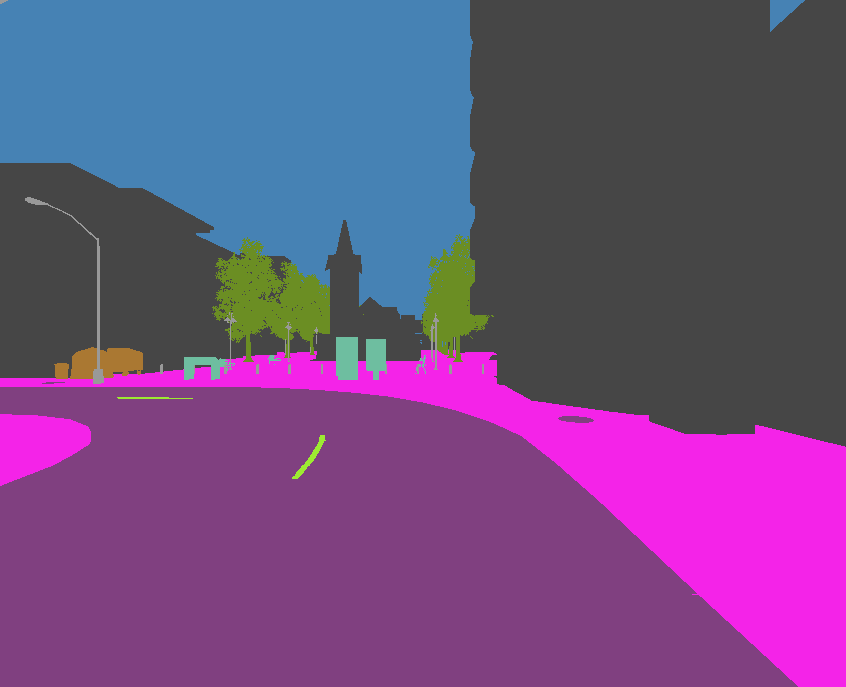} &\hspace{-0.47cm}
\includegraphics[width=0.162\linewidth, height=0.11\linewidth]{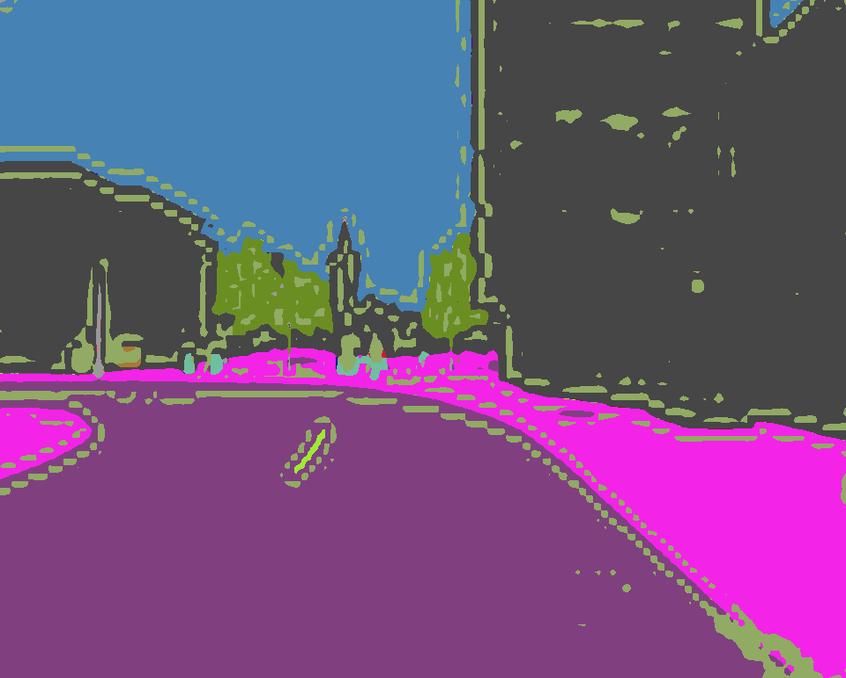} &\hspace{-0.47cm}
\includegraphics[width=0.162\linewidth, height=0.11\linewidth]{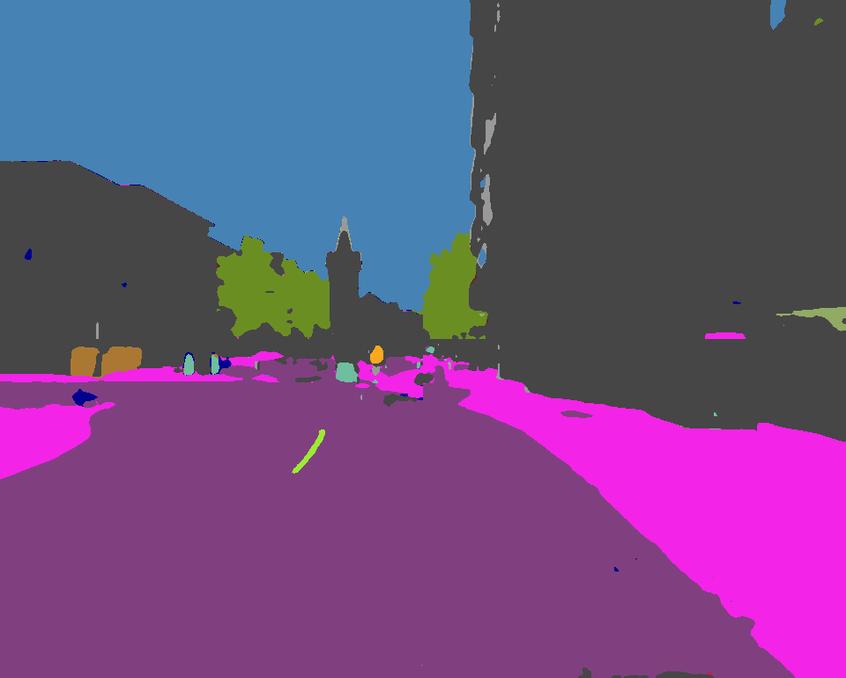} &\hspace{-0.47cm}
\includegraphics[width=0.162\linewidth, height=0.11\linewidth]{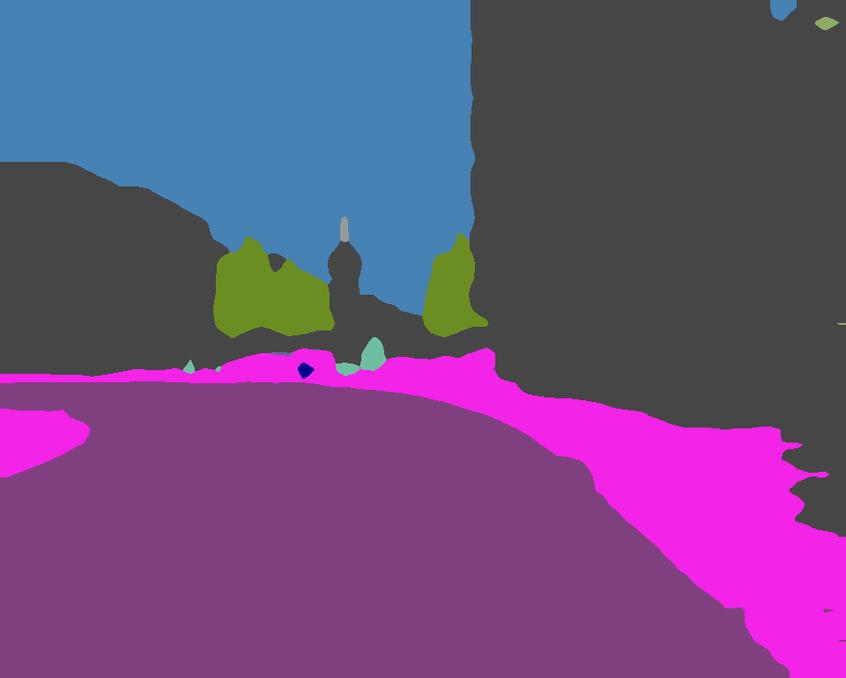} &\hspace{-0.43cm}
\includegraphics[width=0.162\linewidth, height=0.11\linewidth]{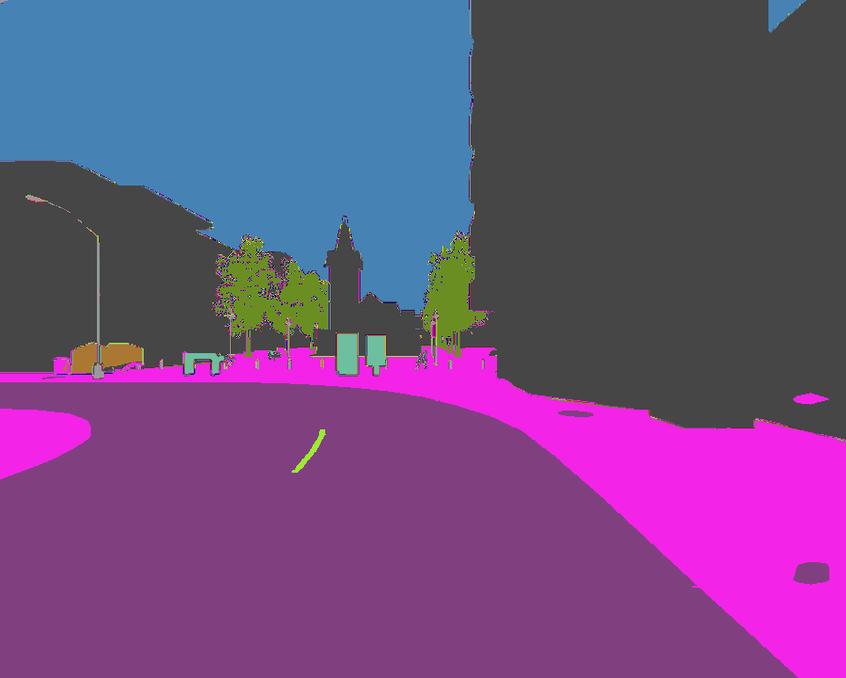}\\
\hline

\includegraphics[width=0.162\linewidth, height=0.11\linewidth]{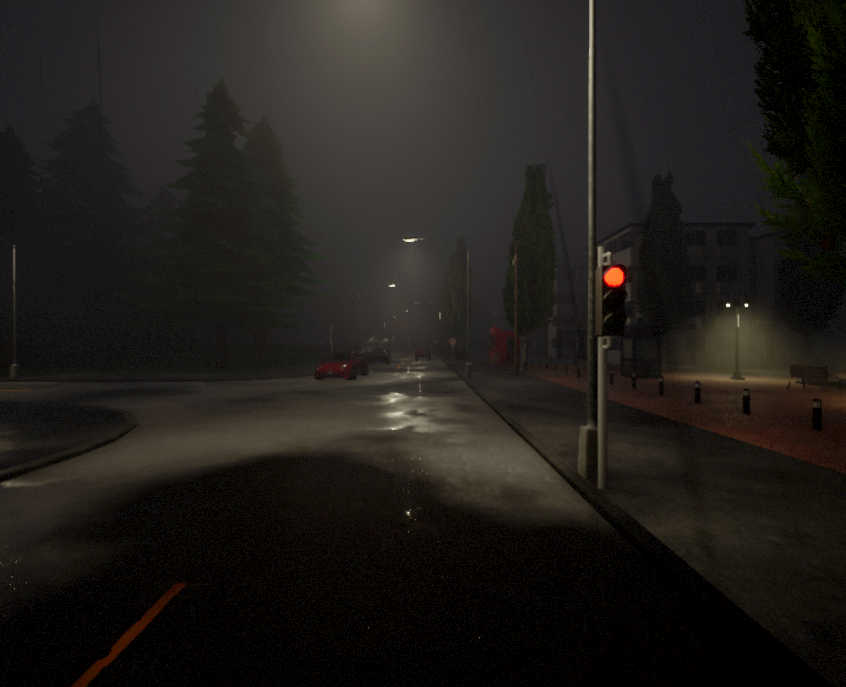} &\hspace{-0.47cm}
\includegraphics[width=0.162\linewidth, height=0.11\linewidth]{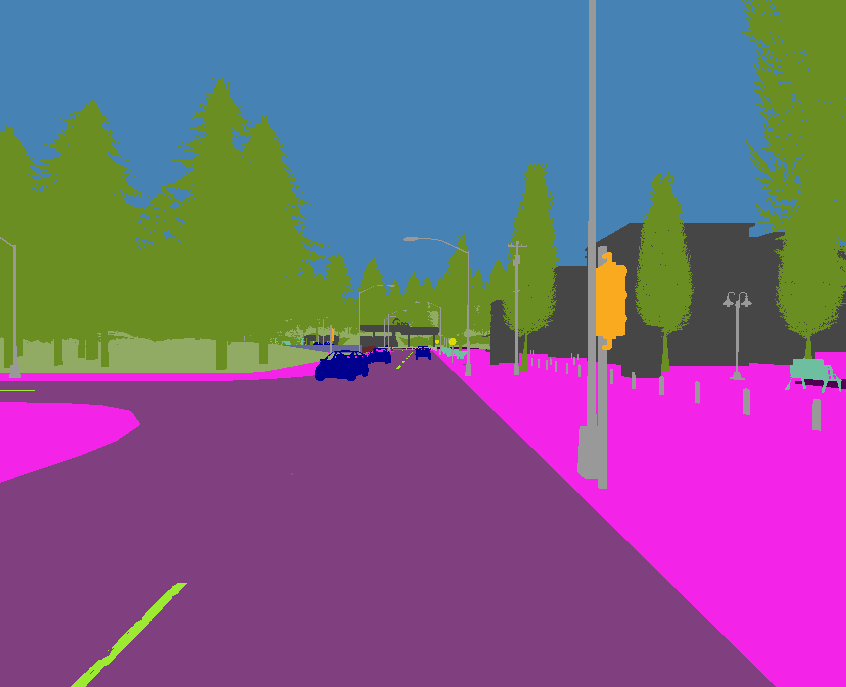} &\hspace{-0.47cm}
\includegraphics[width=0.162\linewidth, height=0.11\linewidth]{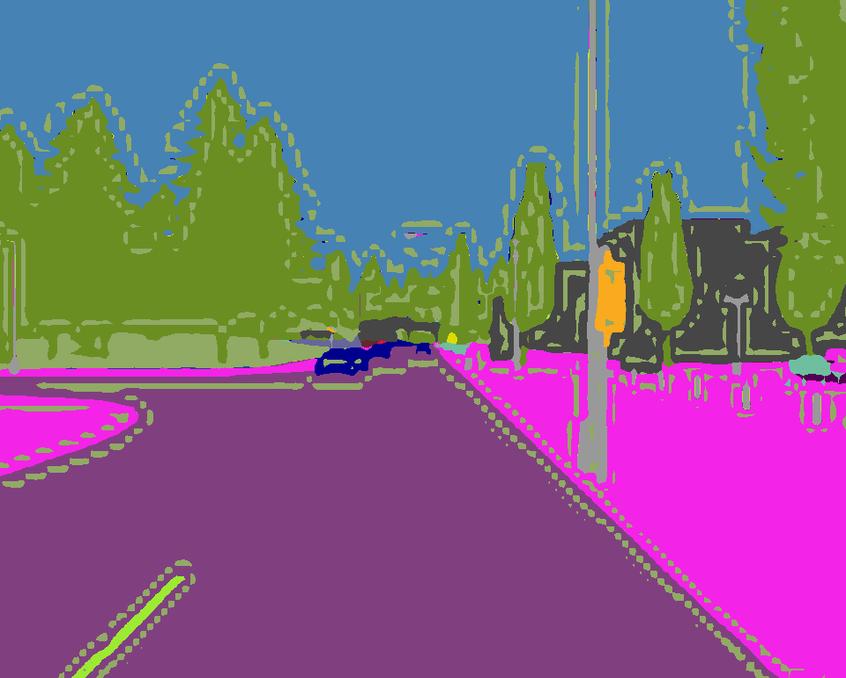} &\hspace{-0.47cm}
\includegraphics[width=0.162\linewidth, height=0.11\linewidth]{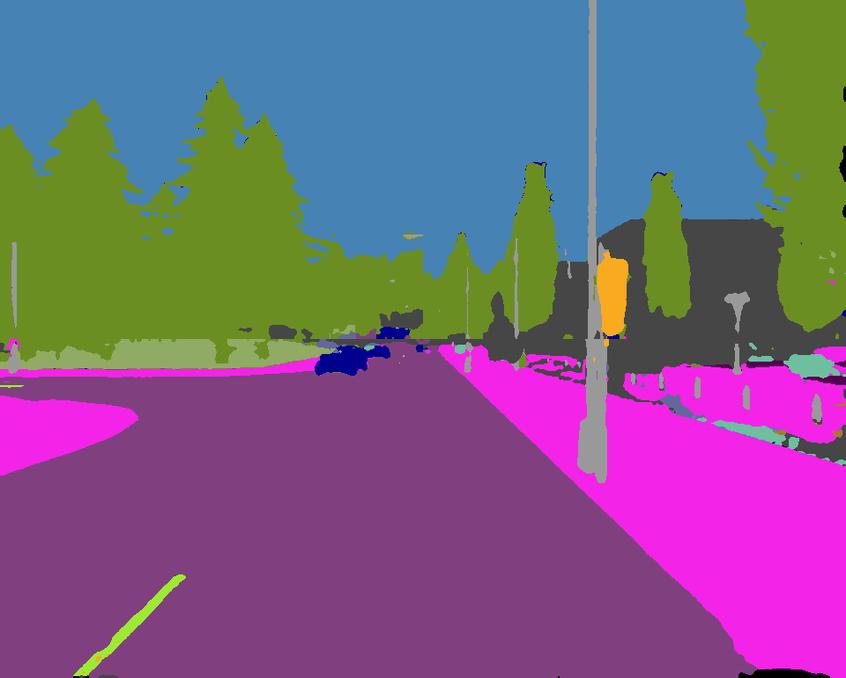} &\hspace{-0.47cm}
\includegraphics[width=0.162\linewidth, height=0.11\linewidth]{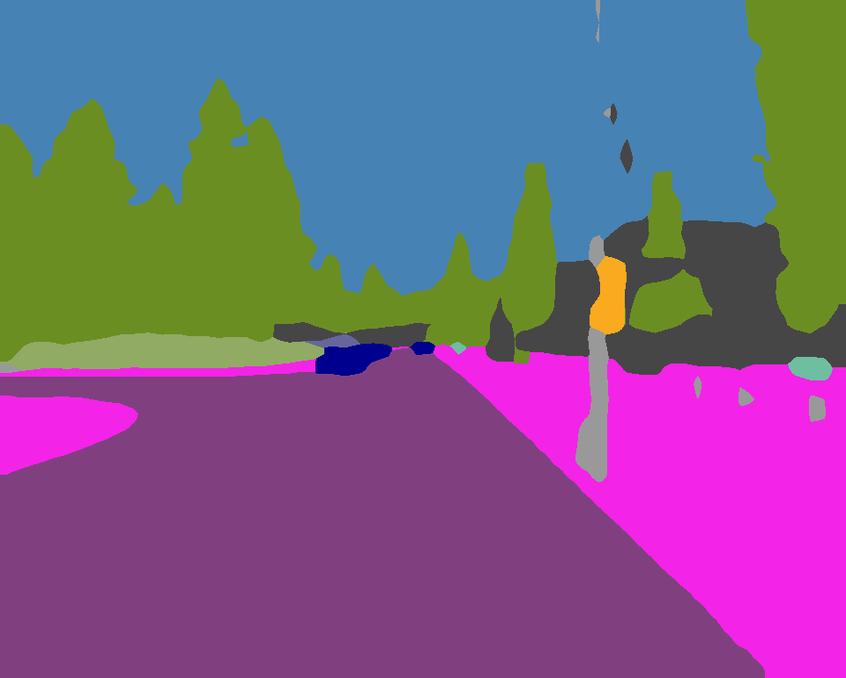} &\hspace{-0.43cm}
\includegraphics[width=0.162\linewidth, height=0.11\linewidth]{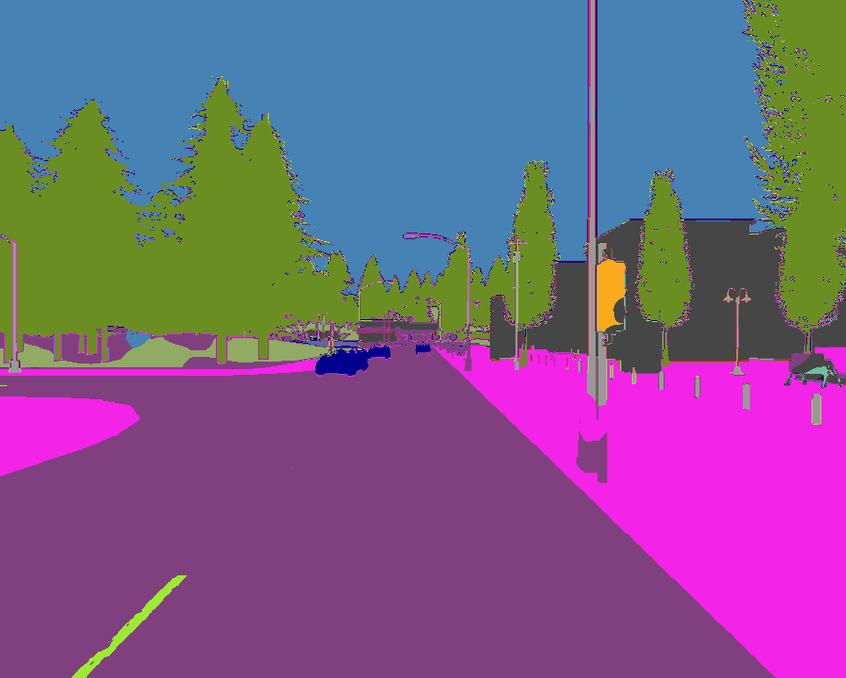}\\
\hline

\includegraphics[width=0.162\linewidth, height=0.11\linewidth]{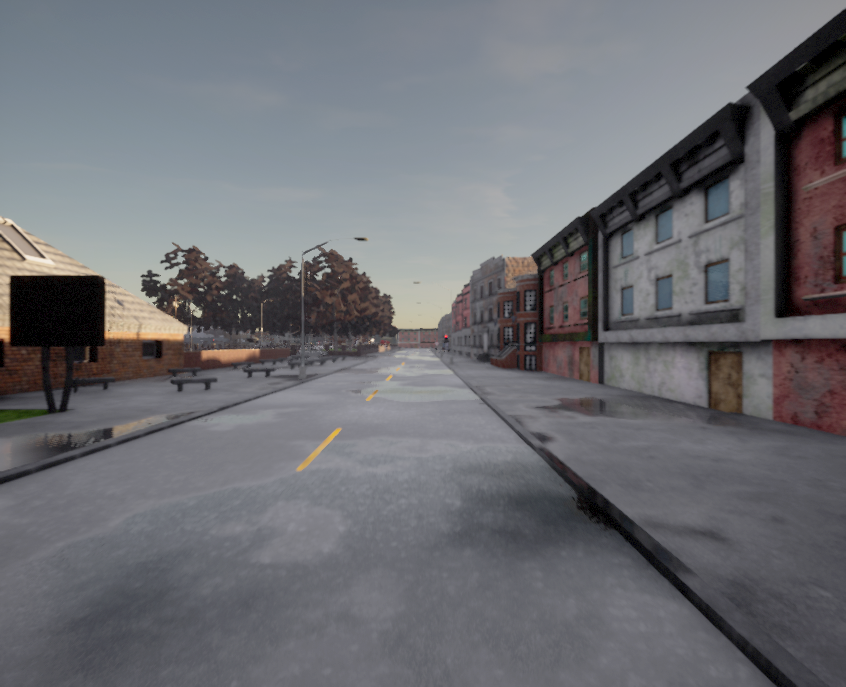} &\hspace{-0.47cm}
\includegraphics[width=0.162\linewidth, height=0.11\linewidth]{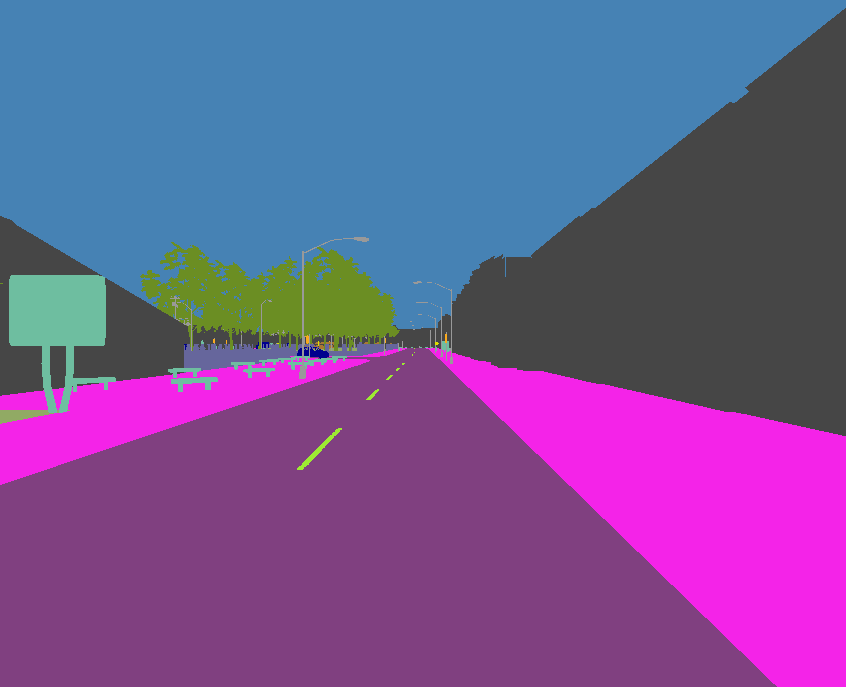} &\hspace{-0.47cm}
\includegraphics[width=0.162\linewidth, height=0.11\linewidth]{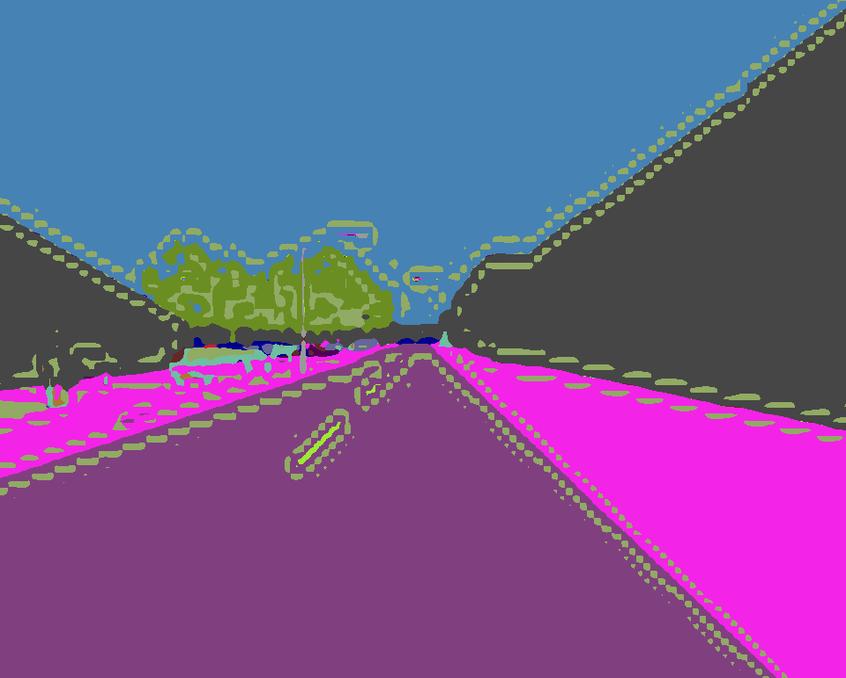} &\hspace{-0.47cm}
\includegraphics[width=0.162\linewidth, height=0.11\linewidth]{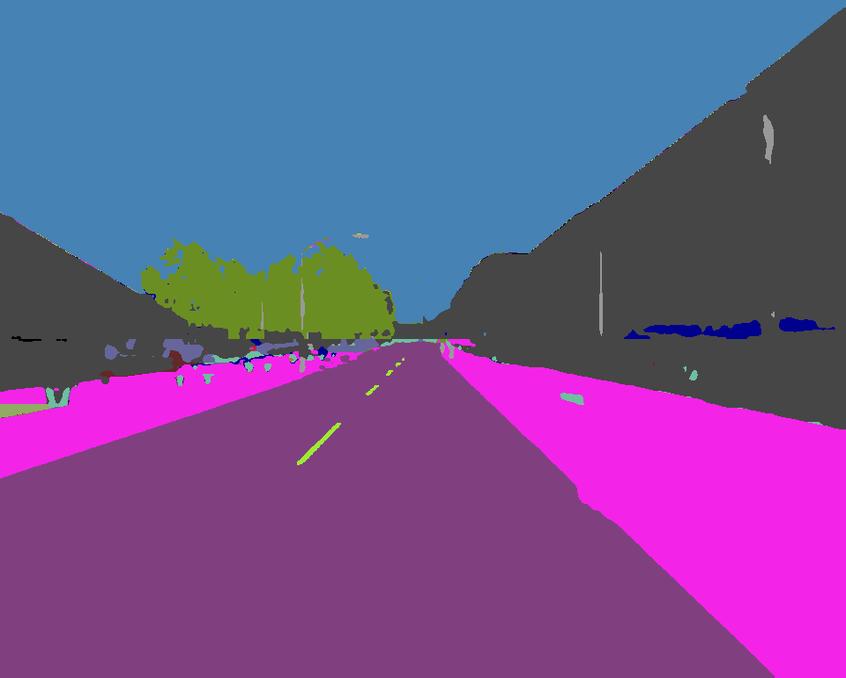} &\hspace{-0.47cm}
\includegraphics[width=0.162\linewidth, height=0.11\linewidth]{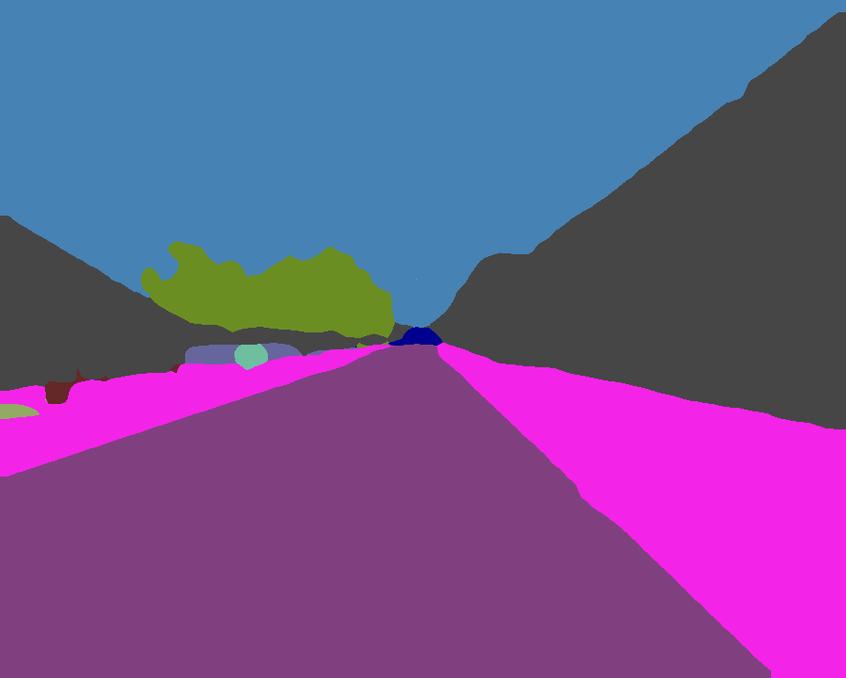} &\hspace{-0.43cm}
\includegraphics[width=0.162\linewidth, height=0.11\linewidth]{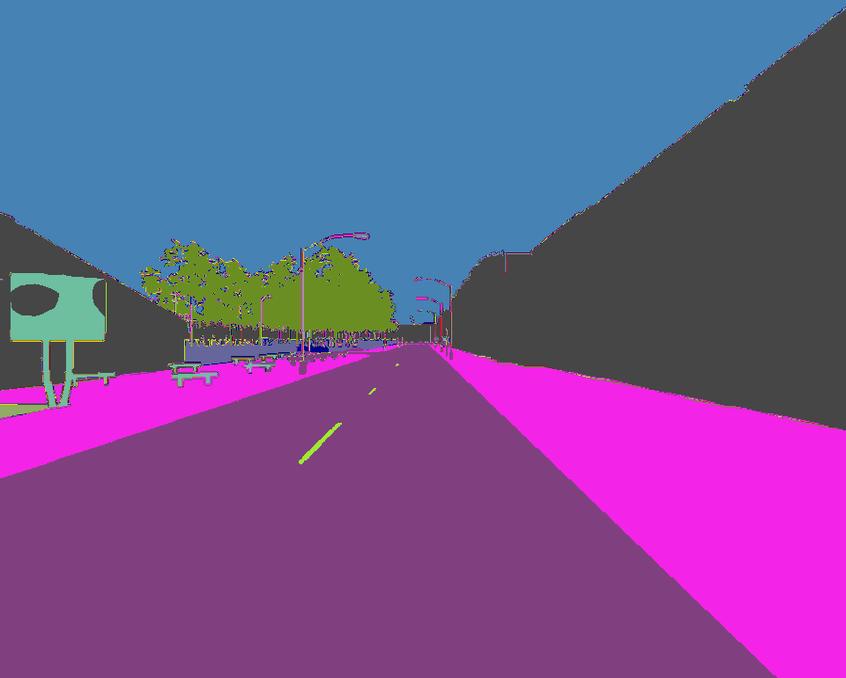}\\
\hline
\end{tabularx}
\caption{Qualitative performance of Advent against other baselines under \textbf{diverse adverse weather conditions}.}
\label{tab:semantic_pred}
\end{table*}

\subsubsection{Unfolding \& Optimization.}
The total loss \( L \) of training is
\begin{align}
    L = L_{CE} + \alpha L_{BD} + \gamma L_{con},
\end{align}
where \( \alpha \) and \( \gamma \) are the weights assigned to \( L_{BD} \) and \( L_{con} \), respectively. In practice, they are tuned based on heuristics or exhaustive searches, which results in instability and suboptimal performance. Treating $\alpha$ and $\beta$ as trainable parameters can avoid these issues.  

To this end, inspired by \cite{monga2021algorithm,10529194}, we propose to treat each optimization iteration as one layer, and treat $\alpha$ and $\beta$ as learnable parameters within each layer. Based on this, above both regularizers can be unfolded into $K$ layers (illustrated in \Cref{Fig.RDU}). Specifically, for layer $k, \text{where}\ k \in \{1, 2, \cdots, K\}$, the output is updated as follow:
\begin{align}
    x^{(k)} =\hspace{0.15cm} &x^{(k-1)}-\eta^{(k)} (\alpha^{(k)}\nabla_x L_{BD}(x^{(k-1)}) \hspace{0.15cm}+\hspace{0.15cm}
    \nonumber \\
    &\gamma^{(k)}\nabla_x L_{con}(x^{(k-1)})),
    \label{Eq:RDU_update}
\end{align}
where $\alpha^{(k)}$ and $\gamma^{(k)}$ are the learnable weights, $\eta^{(k)}$ is the learnable step size, $\nabla_x L_{BD}(x^{(k-1)})$ and $\nabla_x L_{con}(x^{(k-1)})$ are the gradients of $L_{BD}$ and $L_{con}$ relative to pixel $x^{(k-1)}$, respectively, and they can be formulated as follows (denoting $x^{(k-1)}$ as $x$ for short):
\begin{align}
\nabla_x L_{BD}(x) &= - \frac{1}{2}\frac{1}{\sum_{x} \sqrt{P_X(x) P_Y(x)}}  \frac{\sqrt{P_Y(x)}}{\sqrt{P_X(x)}},
\label{Eq:grad_L_BD}
\\
\nabla_x L_{\text{con}}(x) &= \nabla_x B / B - \nabla_x A / A,
\label{Eq:grad_L_con}
\end{align}
where
\begin{align}
A &= \sum\nolimits_{m=1}^{|\mathcal{U}|} \exp(\sigma(x, u_{m}) / \tau),
\label{Eq:L_con_A}
\\
B &= A + \sum\nolimits_{n=1}^{|\mathcal{V}|} \exp(\sigma(x, v_{n}) / \tau),
\label{Eq:L_con_B}
\\
\nabla_x A &= \frac{1}{\tau} \sum\nolimits_{m=1}^{|\mathcal{U}|} \exp(\sigma(x, u_{m}) / \tau) \cdot u_{m}, 
\label{Eq:grad_L_con_A}
\\
\!\nabla_x B \! &=\! \nabla_x A\! +\! \frac{1}{\tau}  \sum\nolimits_{n=1}^{|\mathcal{V}|}\! \exp(\sigma(x, v_{n}) / \tau) \cdot v_{n}.
\label{Eq:grad_L_con_B}
\end{align}

After $K$ steps of gradient descent updates by $K$ cascading layers, the output of the $K$-th layer is regularized by both regularizers, and then is added to the either-or output of LSM. Then we use the summation results serving as prediction and the ground truth to calculate $L_{CE}$ to optimize involved backbone(s), head and $\{\alpha^{(k)}$, $\gamma^{(k)}$, $\eta^{(k)}\}_{k=1}^K$ by back propagation. In conclusion, the unfolding and optimization of URs are illustrated in \Cref{Fig.RDU}, and we also algorithmize the unfolding and optimization process of URs in Algorithm \ref{Algo:DUN}.

\section{Experiments} \label{sec_experiments}
\subsection{Datasets, Metrics and Implementation}
\subsubsection{Datasets.}
We conduct experiments based on two datasets: the Apolloscapes dataset \cite{wang2019apolloscape} and the CARLA\_ADV dataset derived from the Carla simulator (version 0.9.13) \cite{dosovitskiy2017carla}. The Apolloscape dataset is a comprehensive AD dataset featuring a variety of driving scenarios captured under various weather conditions. Each annotated image has pixel-level labels across 23 classes such as vehicles and pedestrians. The CARLA\_ADV dataset is specifically designed to encompass a range of adverse weather conditions including fog, clouds, rain, darkness, etc., and combinations thereof. It does also contain 23 classes in pixel-level, including vehicle, building, tree, etc.

\subsubsection{Metrics.}
We evaluate the proposed Advent on semantic segmentation task by employing four metrics: mean Intersection over Union (\textbf{mIoU}), mean Precision (\textbf{mPre}), mean Recall (\textbf{mRec}), and mean F1 (\textbf{mF1}). These metrics are defined in \textit{Appendix II of Supplementary Materials}.

\subsubsection{Implementation Details.}
Our involved backbone(s) and head are all implemented using Pytorch on two NVIDIA GeForce 4090 GPUs. For the architecture, we implement the proposed Advent based on ResNet backbone \cite{He_2016_CVPR} and ASSP head \cite{chen2018encoderdecoder}. We select the Adam optimizer for training, configuring it with Betas values at (0.9, 0.999) and the weight decay at 1e-4. The training was executed with a batch size of 8 and a learning rate of 3e-4. In addition, we compare the proposed Advent with other 8 baselines, \ie, DeepLabv3+ \cite{chen2018encoderdecoder}, BiSeNetV2 \cite{yu2021bisenet}, SegNet \cite{badrinarayanan2017segnet}, TopFormer \cite{zhang2022topformer}, SeaFormer \cite{wan2023seaformer}, BASeg \cite{xiao2023baseg}, HRDA \cite{hoyer2023domain}, and AttaNet \cite{song2021attanet}. All these baselines are also implemented using Pytorch. We train Advent and such baselines based on adopted datasets from scratch. The training hardware and software are listed in \textit{Appendix III of Supplementary Materials}.

\begin{table}[t]
\setlength{\tabcolsep}{20.5pt}
\begin{tabularx}{\linewidth}{ccc}
\hline
Methods    & GFLOPs & FPS \\ \hline
DeepLabv3+ & 50.58     & 297.06    \\
BiSeNetV2  & 36.46     & 545.84    \\
SegNet     & 327.93    & 65.54    \\
AttaNet    & 23.91     & 506.40    \\
BASeg      & 562.31    & 35.60    \\
HRDA       & 823.57    & 9.69     \\
SeaFormer  & 13.77     & 234.00    \\
TopFormer  & 3.56      & 490.38    \\
\textbf{Advent (Ours)}    & 50.02     & 28.11     \\ \hline
\end{tabularx}
\caption{The comparison of GFLOPs and FPS of Advent against other baselines.}
\label{Tab:flops_and_fps}
\end{table}

\begin{table}[t]
\setlength{\tabcolsep}{1.8pt}
\begin{tabularx}{\linewidth}{ccccc}
\hline
Depth & mIoU       & mPre       & mRec       & mF1        \\ \hline
Depth-1   & 41.52$\pm$0.77 & \textbf{57.83$\pm$0.92} & 46.90$\pm$0.67  & 49.96$\pm$0.78 \\
Depth-2   & 41.29$\pm$0.80  & 57.43$\pm$0.60  & 46.68$\pm$0.84 & 49.76$\pm$0.80  \\
Depth-3   & 41.58$\pm$0.78 & 57.52$\pm$0.85 & \textbf{47.08$\pm$0.72} & \textbf{50.09$\pm$0.75} \\
Depth-4   & \textbf{41.81$\pm$0.88} & 56.78$\pm$0.70  & 46.23$\pm$0.81 & 49.13$\pm$0.84 \\ \hline
\end{tabularx}
\caption{The effect of LSM depth on Advent performance.}
\label{Tab.LSM_depth_comp}
\end{table}

\subsection{Main Results and Empirical Analyses}
\subsubsection{Quantitative Comparison.}
\Cref{Tab.perf_comp} presents a quantitative performance comparison of Advent with DeepLabv3+, BiSeNetV2, SegNet, AttaNet, BASeg, HRDA, SeaFormer, and TopFormer across Apolloscapes dataset and CARLA\_ADV dataset. In this table, the best-performing model for each metric is highlighted in bold, while the second-best model is underlined. As shown in \Cref{Tab.perf_comp}, Advent generally achieves superior performance compared to all other baselines across nearly all metrics for the adopted datasets. For instance, on the Apolloscapes dataset, Advent outperforms DeepLabv3+, BiSeNetV2, SegNet, AttaNet, BASeg, HRDA, SeaFormer, and TopFormer in mIoU by (59.35 - 26.58) / 26.58 = 123.29\%, 158.94\%, 182.48\%, 188.25\%, 194.69\%, 175.41\%, 191.79\%, and 190.79\%, respectively. These results show the superiority of Advent over other baselines.

\begin{figure}[t]
\hspace{-0.2cm}
\subfloat[\small mIoU]{
\label{Fig.LSM_depth_mIoU}
\includegraphics[width=0.48\linewidth, height=0.35\linewidth]{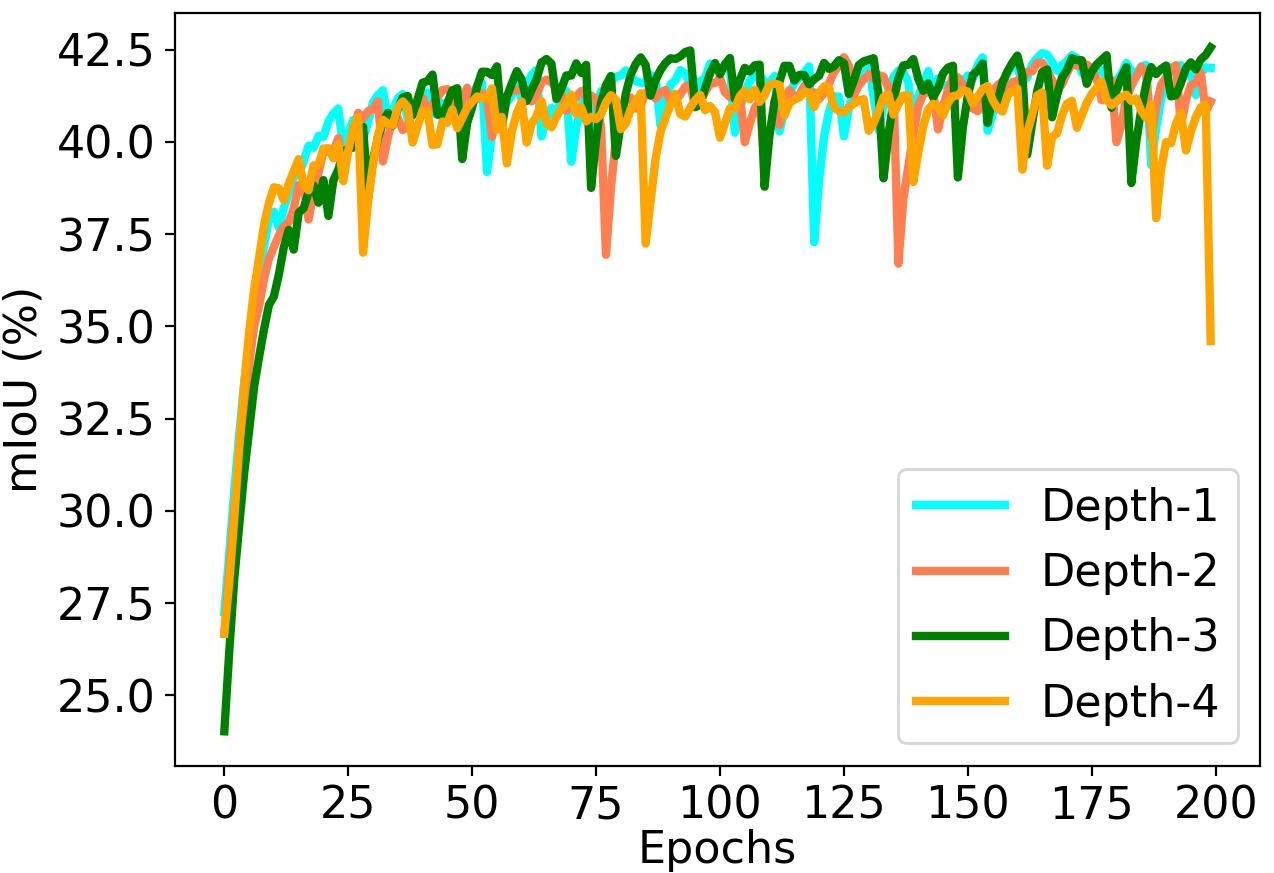}}
\hspace{-0.1cm}
\subfloat[\small mF1]{
\label{Fig.LSM_depth_mF1}
\includegraphics[width=0.46\linewidth, height=0.35\linewidth]{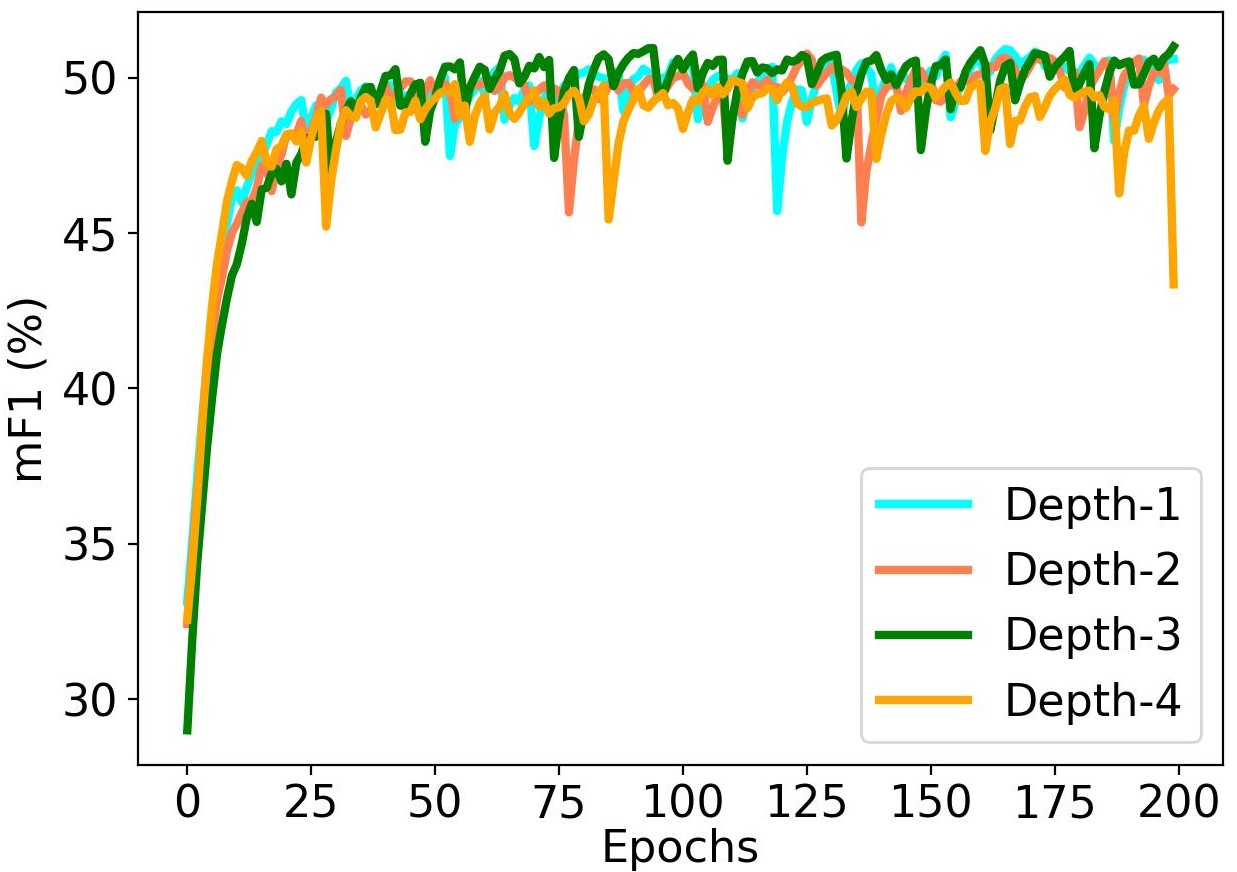}}
\caption{The comparison among different LSM depths.}
\label{Fig.LSM_depth_comp}
\end{figure}

\begin{table}[t]
\setlength{\tabcolsep}{1.9pt}
\begin{tabular}{ccccc}
\hline
\begin{tabular}[c]{@{}c@{}}Merge\\ Policy\end{tabular} & mIoU  & mPre  & mRec  & mF1   \\ \hline
CE                                                     & 42.75$\pm$0.92 & 60.26$\pm$0.66 & 48.32$\pm$0.80 & 51.29$\pm$0.88 \\
FI                                                     & \textbf{43.57$\pm$0.84} & \textbf{60.29$\pm$0.67} & \textbf{49.18$\pm$0.73} & \textbf{52.04$\pm$0.87} \\ \hline
\end{tabular}
\caption{The performance comparison between CE and FI.}
\label{Tab.CE_FI_perf}
\end{table}

\subsubsection{Qualitative Comparison.}
\Cref{tab:semantic_pred} qualitatively compares the inference performance of the proposed Advent with other models under various adverse weather conditions, including rainy, dark, foggy, cloudy, and more. From \Cref{tab:semantic_pred}, the following observations can be made: \textbf{(I)} The proposed Advent demonstrates superior performance across different adverse weather conditions, with predictions closely approximating the ground truth. For instance, in the rainy condition (the third row), Advent's prediction aligns well with the ground truth, whereas other models show various shortcomings, such as class confusion (\eg, HRDA), boundary uncertainty (\eg, DeepLabv3+), and small object errors (\eg, BASeg). \textbf{(II)} Certain models (\eg, HRDA) perform well in clear weather conditions (\eg, the first row) but experience a significant drop in performance under adverse weather conditions (\eg, the third row).

\subsubsection{Complexity and Inference Delay.}
\Cref{Tab:flops_and_fps} presents a comparison of GFLOPs and FPS between Advent and other baseline models. The following insights can be drawn: \textbf{(I)} Models such as TopFormer and BiSeNetV2 exhibit high FPS with low computational requirements, making them well-suited for real-time applications. \textbf{(II)} HRDA and BASeg have the highest GFLOPs (823.57 and 562.31, respectively) but suffer from low FPS, highlighting their computational intensity and slower processing speeds. \textbf{(III)} Advent has a GFLOPs comparable to DeepLabv3+ but demonstrates a significantly lower FPS (28.11) due to its use of heavy input processing and unfolding regularizers.

\subsection{Ablation Study}
This part reveals four types of ablation study: (I) how LSM depth affects Advent's prediction; (II) how CE and FI impact Advent's prediction; (III) how GSM influences Advent's prediction; and (IV) how URs impact Advent's prediction. 

\begin{table}[t]
\setlength{\tabcolsep}{2.4pt}
\begin{tabularx}{\linewidth}{ccccc}
\hline
GSM          & mIoU       & mPre       & mRec       & mF1        \\ \hline
\xmark & 27.12$\pm$1.84 & 46.81$\pm$1.76 & 32.96$\pm$1.69 & 35.36$\pm$1.94 \\
\checkmark  & \textbf{41.52$\pm$0.77} & \textbf{57.83$\pm$0.92} & \textbf{46.90$\pm$0.67} & \textbf{49.96$\pm$0.78} \\  \hline
\end{tabularx}
\caption{The effect of GSM on Advent performance.}
\label{Tab.GSM_comp}
\end{table}

\subsubsection{The Impact of LSM Depth.}
\Cref{Tab.LSM_depth_comp} explores the effect of the LSM depth on performance. Across different LSM depths (Depth-1 through Depth-4), the performance metrics show certain variation, indicating that changes in depth have impact on Advent's performance. \Cref{Fig.LSM_depth_comp} corroborates this finding, showing that all depths reach different performance levels. The variations in performance across different depths indicate that both too shallow (Depth-1) and too deep (Depth-4) configurations do not consistently yield the best results. Depths-3 generally show more stable and slightly improved performances, suggesting that moderate depths could be optimal. This implies that an intermediate LSM depth could potentially offer a better compromise between computation complexity and prediction accuracy.

\begin{figure}[t]
\hspace{-0.4cm}
\subfloat[\small mIoU]{
\label{Fig.GSM_mIoU}
\includegraphics[width=0.49\linewidth, height=0.35\linewidth]{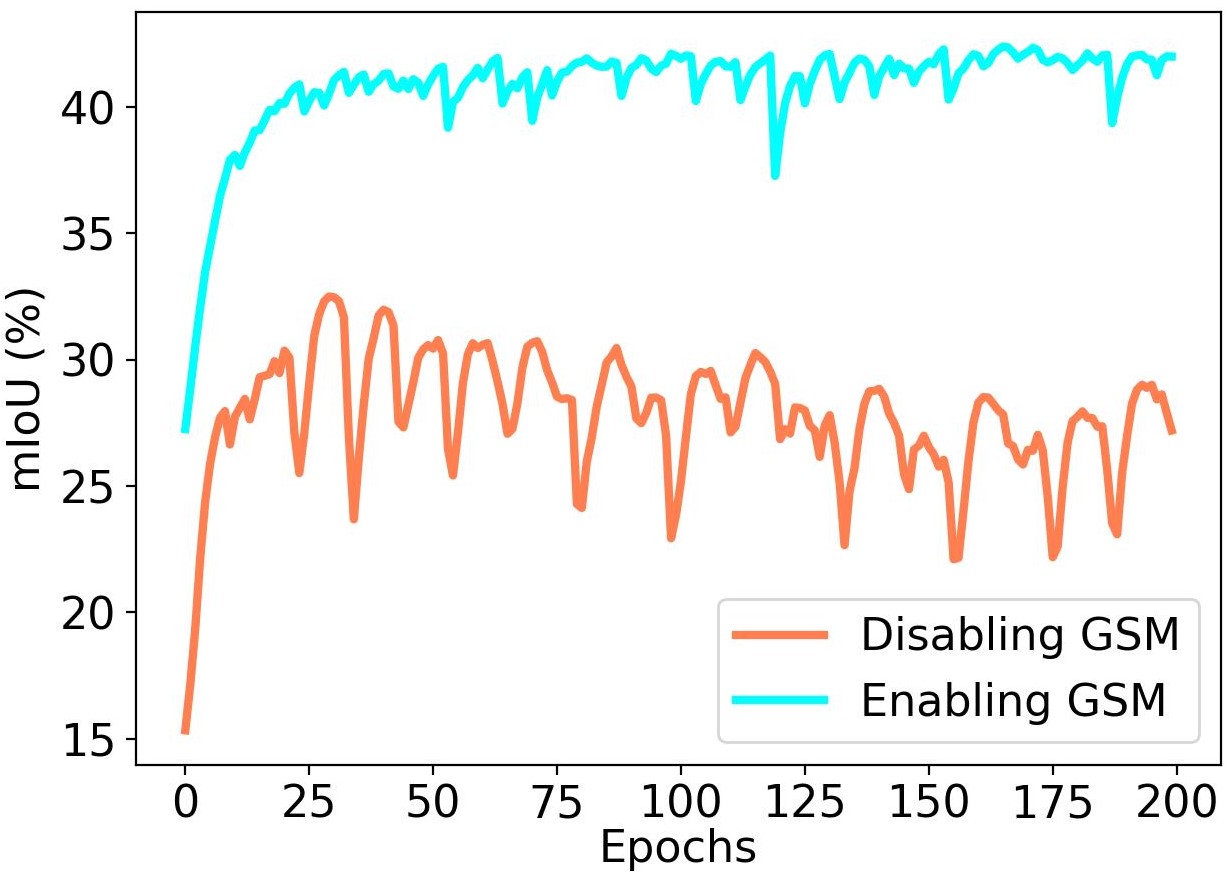}}
\hspace{-0.3cm}
\subfloat[\small mF1]{
\label{Fig.GSM_mF1}
\includegraphics[width=0.49\linewidth, height=0.35\linewidth]{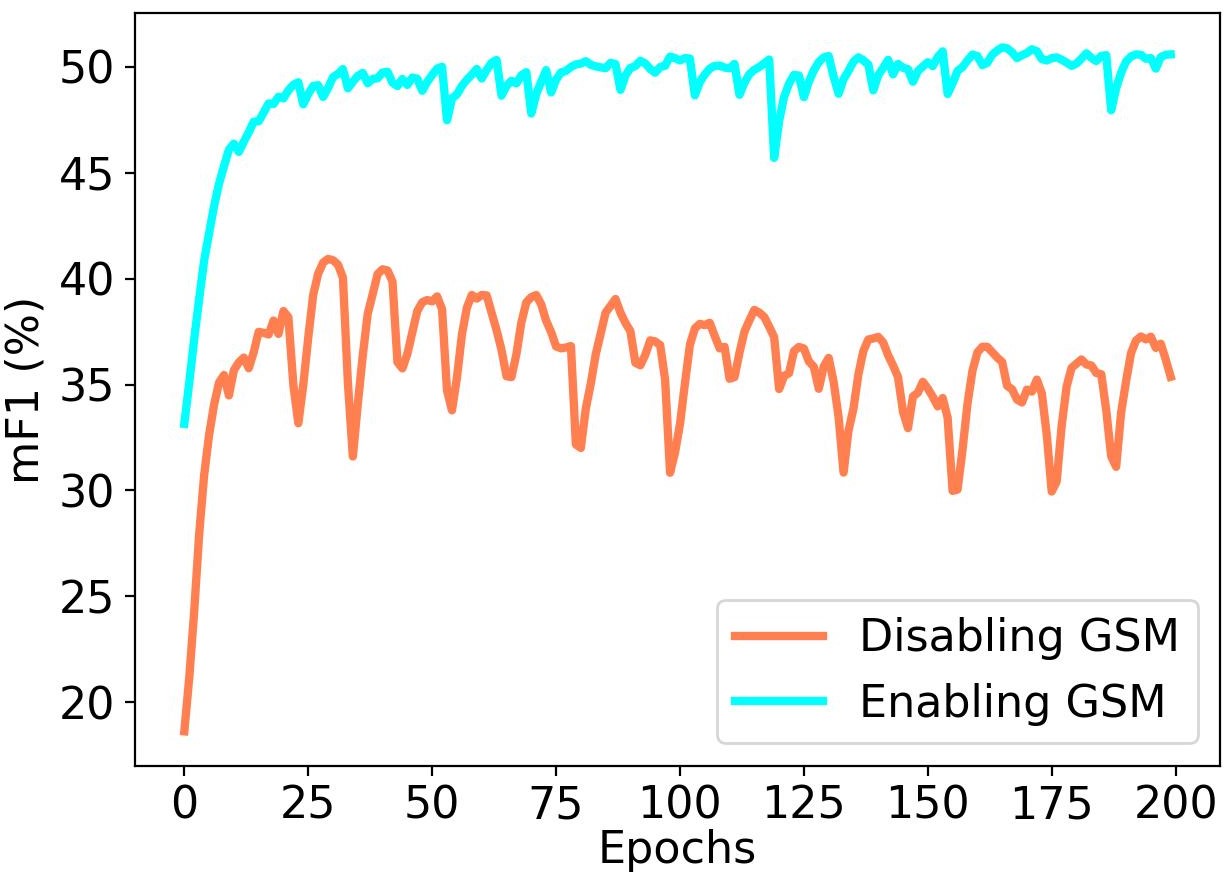}}
\caption{The performance comparison between enabling and disabling GSM.}
\label{Fig.GSM_comp}
\end{figure}

\subsubsection{The Impact of CE and FI.}
\Cref{Fig.EF_vs_LF} illustrates the architecture of CE and FI.
\Cref{Tab.CE_FI_perf} compares the performance of CE and FI across four metrics, where FI performs better than CE in all adopted metrics. Overall, although FI generally outperforms CE, FI has more learnable parameters and thus requires more training time. Therefore, we should take both model performance and training cost into account when we choose which one fits our needs.

\subsubsection{The Impact of GSM.}
\Cref{Tab.GSM_comp} demonstrates that enabling GSM significantly improves the model performance across all measured metrics. \Cref{Fig.GSM_comp} shows that with GSM enabled, the performance metrics are consistently higher and more stable across 200 epochs, compared to the lower and more fluctuating metrics when GSM is disabled. Overall, enabling GSM leads to better performance.

\begin{table}[tp]
\setlength{\tabcolsep}{2.0pt}
\begin{tabularx}{\linewidth}{ccccc}
\hline
Loss                    & mIoU       & mPre       & mRec       & mF1        \\ \hline
CE            & 26.77$\pm$0.77 & 38.35$\pm$0.78 & 31.09$\pm$0.64 & 32.54$\pm$0.87 \\
VRs & 31.49$\pm$0.34 & 50.23$\pm$0.89 & 35.56$\pm$0.40 & 37.84$\pm$0.43 \\
URs ($2$) & 52.13$\pm$2.94 & 68.06$\pm$1.21 & 53.96$\pm$2.86 & 58.02$\pm$2.57 \\
URs ($5$) & \textbf{59.35$\pm$0.57} & \textbf{81.38$\pm$2.22} & \textbf{60.76$\pm$0.49} & \textbf{65.28$\pm$0.55} \\ \hline
\end{tabularx}
\caption{The regularizer comparison.}
\label{Tab.regularizer_comp}
\end{table}

\begin{figure}[!htbp]
\hspace{-0.4cm}
\subfloat[\small mIoU]{
\label{Fig.regularizer_mIoU}
\includegraphics[width=0.50\linewidth, height=0.35\linewidth]{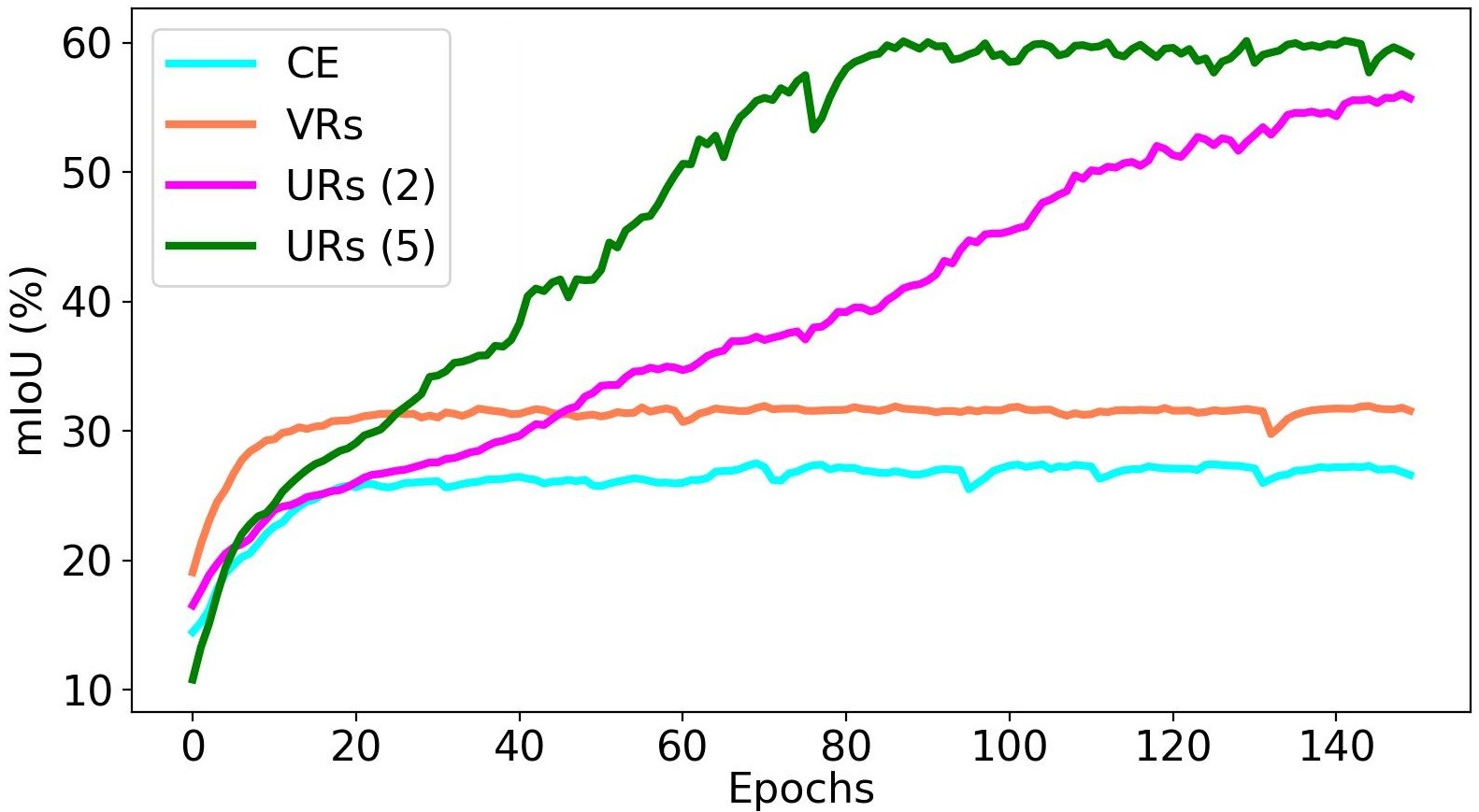}}
\hspace{-0.1cm}
\subfloat[\small mF1]{
\label{Fig.regularizer_mF1}
\includegraphics[width=0.49\linewidth, height=0.35\linewidth]{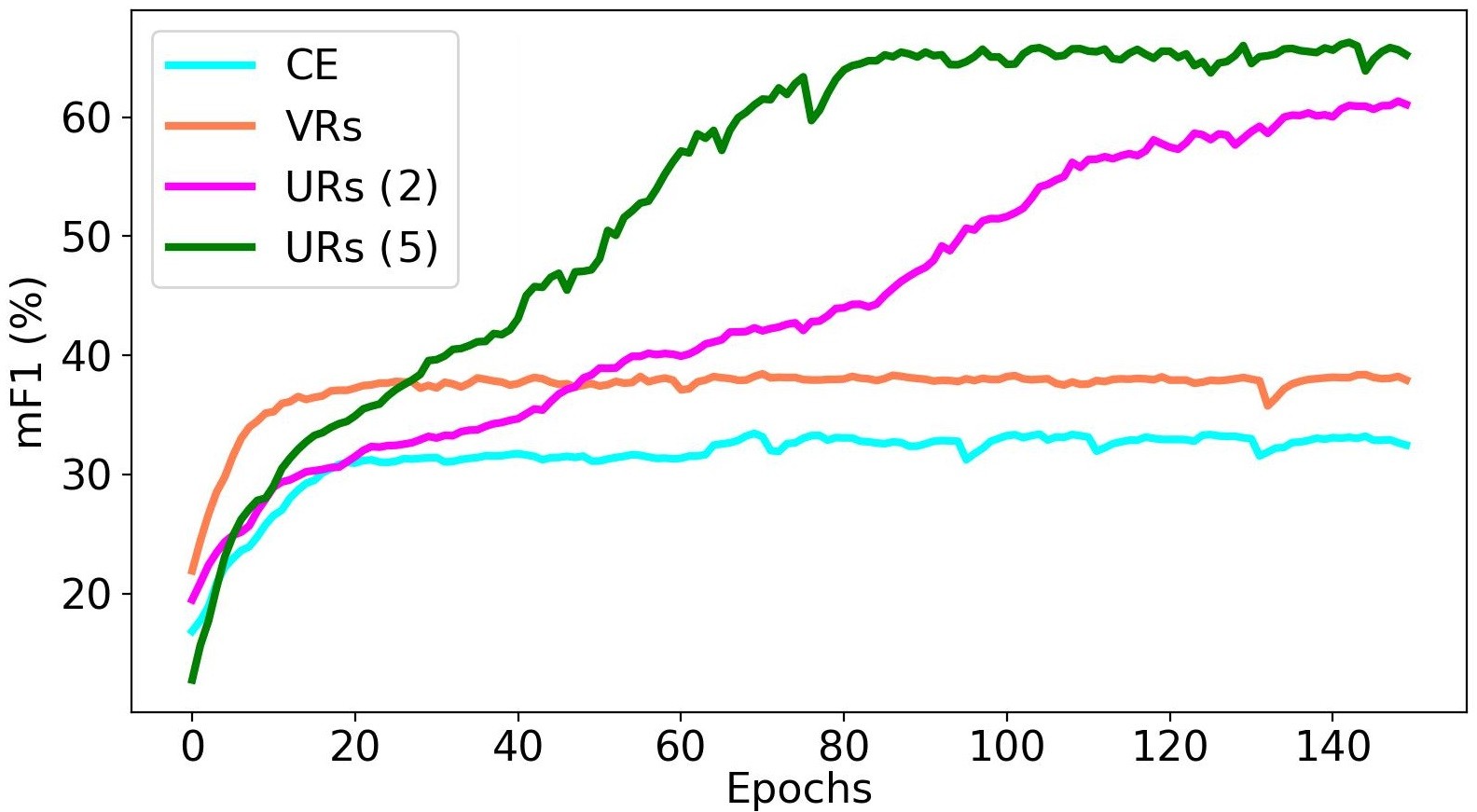}}
\caption{The regularizer comparison.}
\label{Fig.regularizer_comp}
\end{figure}

\subsubsection{The Impact of URs.}
\Cref{Tab.regularizer_comp} and \Cref{Fig.regularizer_comp} compare the performance of the following cases: (I) only cross entropy loss (denoted as CE); (II) cross entropy plus vanilla regularizers (denoted as VRs); (III) cross entropy plus 2-layer unfolded regularizers (denoted as URs ($2$)); (IV) cross entropy plus 5-layer unfolded regularizers (denoted as URs ($5$)). By comparing these cases, we can observe the following patterns: (I) \Cref{Tab.regularizer_comp} shows a consistent improvement in performance when additional regularization techniques are added to the basic CE loss, leading to better scores in mIoU, mPre, mRec, and mF1. (II) \Cref{Tab.regularizer_comp} also demonstrates that URs outperform corresponding VRs in almost all metrics. For example, URs ($2$) and URs ($5$) achieve mIoUs of 52.13 and 59.35, respectively, compared to 31.49 achieved by VRs. (III) \Cref{Tab.regularizer_comp} additionally shows that the more layers the regularizers are unfolded, the better the performance is obtained. For instance, URs ($5$) achieve an mIoU of 59.35, better than the 52.13 achieved by URs ($2$). These patterns can be visually confirmed in \Cref{Fig.regularizer_comp}. In addition, \Cref{Fig.regularizer_comp} illustrates that URs ($5$) converges faster than URs ($2$). 

\section{Conclusion}
AD perception in various adverse weather conditions is challenging. This work proposed Advent to enhance the model generalization across the mixture of multiple adverse weather conditions by considering local temporal correlation via LSM, global randomness via GSM, and unfolded regularization via URs. LSM processed frame sequences using InsU, IntU and DU to capture immediate scenes, consistent background and dynamic changes. Then, GSM shuffles these LSM-processed segments from different positions in input sequence to avoid LSM-induced overfitting. In addition, Advent also contains an unfolded image-level regularizer and an unfolded inter-class contrastive regularizer to enhance the prediction generalization across the mixture of multiple adverse weather conditions. Extensive experimental results show that Advent exhibits impressive generalization and outperforms existing methods significantly.


\begin{thebibliography}{50}
\providecommand{\natexlab}[1]{#1}

\bibitem[{Alzanin(2025)}]{alzanin2025explainable}
Alzanin, S. 2025.
\newblock Explainable artificial intelligence with temporal convolutional networks for adverse weather condition detection in driverless vehicles.
\newblock \emph{Scientific Reports}, 15(1): 19475.

\bibitem[{Badrinarayanan, Kendall, and Cipolla(2017)}]{badrinarayanan2017segnet}
Badrinarayanan, V.; Kendall, A.; and Cipolla, R. 2017.
\newblock Segnet: A deep convolutional encoder-decoder architecture for image segmentation.
\newblock \emph{IEEE transactions on pattern analysis and machine intelligence}, 39(12): 2481--2495.

\bibitem[{Bhattacharyya(1943)}]{bhattacharyya1943measure}
Bhattacharyya, A. 1943.
\newblock On a measure of divergence between two statistical populations defined by their probability distribution.
\newblock \emph{Bulletin of the Calcutta Mathematical Society}, 35: 99--110.

\bibitem[{Bi, You, and Gevers(2024)}]{bi2024learning}
Bi, Q.; You, S.; and Gevers, T. 2024.
\newblock Learning generalized segmentation for foggy-scenes by bi-directional wavelet guidance.
\newblock In \emph{Proceedings of the AAAI Conference on Artificial Intelligence}, volume~38, 801--809.

\bibitem[{Chen et~al.(2018)Chen, Zhu, Papandreou, Schroff, and Adam}]{chen2018encoderdecoder}
Chen, L.-C.; Zhu, Y.; Papandreou, G.; Schroff, F.; and Adam, H. 2018.
\newblock Encoder-Decoder with Atrous Separable Convolution for Semantic Image Segmentation.
\newblock arXiv:1802.02611.

\bibitem[{Chen et~al.(2024)Chen, Ma, Qiao, and Wang}]{chen2024m}
Chen, S.; Ma, Y.; Qiao, Y.; and Wang, Y. 2024.
\newblock M-bev: Masked bev perception for robust autonomous driving.
\newblock In \emph{Proceedings of the AAAI Conference on Artificial Intelligence}, volume~38, 1183--1191.

\bibitem[{Dosovitskiy et~al.(2017)Dosovitskiy, Ros, Codevilla, Lopez, and Koltun}]{dosovitskiy2017carla}
Dosovitskiy, A.; Ros, G.; Codevilla, F.; Lopez, A.; and Koltun, V. 2017.
\newblock CARLA: An open urban driving simulator.
\newblock In \emph{Proceedings of The 1st Annual Conference on Robot Learning}, 1--16.

\bibitem[{Fu et~al.(2024)Fu, Chang, Ling, Zhang, and Qi}]{fu2024auxiliary}
Fu, Z.; Chang, K.; Ling, M.; Zhang, Q.; and Qi, E. 2024.
\newblock Auxiliary Domain-guided Adaptive Detection in Adverse Weather Conditions.
\newblock In \emph{Proceedings of the Asian Conference on Computer Vision}, 3964--3981.

\bibitem[{Gao et~al.(2024)Gao, Li, Salzmann, and He}]{gao2024generalize}
Gao, Z.; Li, B.; Salzmann, M.; and He, X. 2024.
\newblock Generalize or detect? towards robust semantic segmentation under multiple distribution shifts.
\newblock \emph{Advances in Neural Information Processing Systems}, 37: 52014--52039.

\bibitem[{Han et~al.(2025)Han, Guo, Xu, and Shen}]{han2025dme}
Han, W.; Guo, D.; Xu, C.-Z.; and Shen, J. 2025.
\newblock Dme-driver: Integrating human decision logic and 3d scene perception in autonomous driving.
\newblock In \emph{Proceedings of the AAAI Conference on Artificial Intelligence}, volume~39, 3347--3355.

\bibitem[{Hao et~al.(2024)Hao, Wei, Yang, Zhao, Zhang, Zhou, Wang, Li, Kong, and Zhang}]{hao2024your}
Hao, X.; Wei, M.; Yang, Y.; Zhao, H.; Zhang, H.; Zhou, Y.; Wang, Q.; Li, W.; Kong, L.; and Zhang, J. 2024.
\newblock Is your hd map constructor reliable under sensor corruptions?
\newblock \emph{Advances in Neural Information Processing Systems}, 37: 22441--22482.

\bibitem[{He et~al.(2016)He, Zhang, Ren, and Sun}]{He_2016_CVPR}
He, K.; Zhang, X.; Ren, S.; and Sun, J. 2016.
\newblock Deep Residual Learning for Image Recognition.
\newblock In \emph{Proceedings of the IEEE Conference on Computer Vision and Pattern Recognition (CVPR)}.

\bibitem[{Hoyer, Dai, and Van~Gool(2024)}]{hoyer2023domain}
Hoyer, L.; Dai, D.-X.; and Van~Gool, L. 2024.
\newblock Domain adaptive and generalizable network architectures and training strategies for semantic image segmentation.
\newblock \emph{IEEE Transactions on Pattern Analysis and Machine Intelligence}.

\bibitem[{Huang et~al.(2024)Huang, Wu, Li, Fan, Wen, and Wang}]{huang2024sunshine}
Huang, X.; Wu, H.; Li, X.; Fan, X.; Wen, C.; and Wang, C. 2024.
\newblock Sunshine to rainstorm: Cross-weather knowledge distillation for robust 3d object detection.
\newblock In \emph{Proceedings of the AAAI Conference on Artificial Intelligence}, volume~38, 2409--2416.

\bibitem[{Jiang et~al.(2024)Jiang, Huang, Xie, Lei, Li, Shao, and Lu}]{jiang2024domain}
Jiang, K.; Huang, J.; Xie, W.; Lei, J.; Li, Y.; Shao, L.; and Lu, S. 2024.
\newblock Domain adaptation for large-vocabulary object detectors.
\newblock \emph{Advances in Neural Information Processing Systems}, 37: 75422--75453.

\bibitem[{Kalb and Beyerer(2023)}]{kalb2023principles}
Kalb, T.; and Beyerer, J. 2023.
\newblock Principles of forgetting in domain-incremental semantic segmentation in adverse weather conditions.
\newblock In \emph{Proceedings of the IEEE/CVF Conference on Computer Vision and Pattern Recognition}, 19508--19518.

\bibitem[{Karim et~al.(2023)Karim, Mithun, Rajvanshi, Chiu, Samarasekera, and Rahnavard}]{karim2023c}
Karim, N.; Mithun, N.~C.; Rajvanshi, A.; Chiu, H.-p.; Samarasekera, S.; and Rahnavard, N. 2023.
\newblock C-sfda: A curriculum learning aided self-training framework for efficient source free domain adaptation.
\newblock In \emph{Proceedings of the IEEE/CVF conference on computer vision and pattern recognition}, 24120--24131.

\bibitem[{Kim et~al.(2024)Kim, Lee, Choe, and Shim}]{kim2024weakly}
Kim, D.; Lee, S.; Choe, J.; and Shim, H. 2024.
\newblock Weakly supervised semantic segmentation for driving scenes.
\newblock In \emph{Proceedings of the AAAI conference on artificial intelligence}, volume~38, 2741--2749.

\bibitem[{Kim and Byun(2020)}]{kim2020learning}
Kim, M.; and Byun, H. 2020.
\newblock Learning texture invariant representation for domain adaptation of semantic segmentation.
\newblock In \emph{Proceedings of the IEEE/CVF conference on computer vision and pattern recognition}, 12975--12984.

\bibitem[{Li et~al.(2024{\natexlab{a}})Li, Li, Tu, Liu, Guo, Juefei-Xu, Xu, and Yu}]{li2024light}
Li, J.; Li, B.; Tu, Z.; Liu, X.; Guo, Q.; Juefei-Xu, F.; Xu, R.; and Yu, H. 2024{\natexlab{a}}.
\newblock Light the night: A multi-condition diffusion framework for unpaired low-light enhancement in autonomous driving.
\newblock In \emph{Proceedings of the IEEE/CVF Conference on Computer Vision and Pattern Recognition}, 15205--15215.

\bibitem[{Li et~al.(2023{\natexlab{a}})Li, Xu, Ma, Zou, Ma, and Yu}]{li2023domain}
Li, J.; Xu, R.; Ma, J.; Zou, Q.; Ma, J.; and Yu, H. 2023{\natexlab{a}}.
\newblock Domain adaptive object detection for autonomous driving under foggy weather.
\newblock In \emph{Proceedings of the IEEE/CVF winter conference on applications of computer vision}, 612--622.

\bibitem[{Li et~al.(2023{\natexlab{b}})Li, Xie, Li, Liu, and Cheng}]{li2023vblc}
Li, M.; Xie, B.; Li, S.; Liu, C.~H.; and Cheng, X. 2023{\natexlab{b}}.
\newblock VBLC: visibility boosting and logit-constraint learning for domain adaptive semantic segmentation under adverse conditions.
\newblock In \emph{Proceedings of the AAAI Conference on Artificial Intelligence}, volume~37, 8605--8613.

\bibitem[{Li et~al.(2024{\natexlab{b}})Li, Yin, Li, Xu, Yang, and Shen}]{li2024di}
Li, X.; Yin, J.; Li, W.; Xu, C.; Yang, R.; and Shen, J. 2024{\natexlab{b}}.
\newblock Di-v2x: Learning domain-invariant representation for vehicle-infrastructure collaborative 3d object detection.
\newblock In \emph{Proceedings of the AAAI Conference on Artificial Intelligence}, volume~38, 3208--3215.

\bibitem[{Lin et~al.(2024{\natexlab{a}})Lin, Li, Kou, Chang, and Wu}]{10529194}
Lin, Q.; Li, Y.; Kou, W.-B.; Chang, T.-H.; and Wu, Y.-C. 2024{\natexlab{a}}.
\newblock Communication-Efficient Activity Detection for Cell-Free Massive MIMO: An Augmented Model-Driven End-to-End Learning Framework.
\newblock \emph{IEEE Transactions on Wireless Communications}, 1--1.

\bibitem[{Lin et~al.(2024{\natexlab{b}})Lin, Wang, Qi, Dong, and Yang}]{lin2024bev}
Lin, Z.; Wang, Y.; Qi, S.; Dong, N.; and Yang, M.-H. 2024{\natexlab{b}}.
\newblock Bev-mae: Bird’s eye view masked autoencoders for point cloud pre-training in autonomous driving scenarios.
\newblock In \emph{Proceedings of the AAAI conference on artificial intelligence}, volume~38, 3531--3539.

\bibitem[{Ma et~al.(2024)Ma, Wang, Chen, and Shen}]{ma2024slowtrack}
Ma, C.; Wang, N.; Chen, Q.~A.; and Shen, C. 2024.
\newblock Slowtrack: Increasing the latency of camera-based perception in autonomous driving using adversarial examples.
\newblock In \emph{Proceedings of the AAAI conference on artificial intelligence}, volume~38, 4062--4070.

\bibitem[{Mirza et~al.(2022)Mirza, Masana, Possegger, and Bischof}]{mirza2022efficient}
Mirza, M.~J.; Masana, M.; Possegger, H.; and Bischof, H. 2022.
\newblock An efficient domain-incremental learning approach to drive in all weather conditions.
\newblock In \emph{Proceedings of the IEEE/CVF conference on computer vision and pattern recognition}, 3001--3011.

\bibitem[{Monga, Li, and Eldar(2021)}]{monga2021algorithm}
Monga, V.; Li, Y.; and Eldar, Y.~C. 2021.
\newblock Algorithm unrolling: Interpretable, efficient deep learning for signal and image processing.
\newblock \emph{IEEE Signal Processing Magazine}, 38(2): 18--44.

\bibitem[{Najibi et~al.(2023)Najibi, Ji, Zhou, Qi, Yan, Ettinger, and Anguelov}]{najibi2023unsupervised}
Najibi, M.; Ji, J.; Zhou, Y.; Qi, C.~R.; Yan, X.; Ettinger, S.; and Anguelov, D. 2023.
\newblock Unsupervised 3d perception with 2d vision-language distillation for autonomous driving.
\newblock In \emph{Proceedings of the IEEE/CVF International Conference on Computer Vision}, 8602--8612.

\bibitem[{Oord, Li, and Vinyals(2018)}]{oord2018representation}
Oord, A. v.~d.; Li, Y.; and Vinyals, O. 2018.
\newblock Representation learning with contrastive predictive coding.
\newblock \emph{arXiv preprint arXiv:1807.03748}.

\bibitem[{Rothmeier, Huber, and Knoll(2024{\natexlab{a}})}]{rothmeier2024time}
Rothmeier, T.; Huber, W.; and Knoll, A.~C. 2024{\natexlab{a}}.
\newblock Time to shine: Fine-tuning object detection models with synthetic adverse weather images.
\newblock In \emph{Proceedings of the IEEE/CVF Winter Conference on Applications of Computer Vision}, 4447--4456.

\bibitem[{Rothmeier, Huber, and Knoll(2024{\natexlab{b}})}]{Rothmeier_2024_WACV}
Rothmeier, T.; Huber, W.; and Knoll, A.~C. 2024{\natexlab{b}}.
\newblock Time To Shine: Fine-Tuning Object Detection Models With Synthetic Adverse Weather Images.
\newblock In \emph{Proceedings of the IEEE/CVF Winter Conference on Applications of Computer Vision (WACV)}, 4447--4456.

\bibitem[{Song, Mei, and Huang(2021)}]{song2021attanet}
Song, Q.; Mei, K.; and Huang, R. 2021.
\newblock AttaNet: Attention-augmented network for fast and accurate scene parsing.
\newblock In \emph{Proceedings of the AAAI conference on artificial intelligence}, volume~35, 2567--2575.

\bibitem[{Tang et~al.(2025)Tang, Zhang, Yang, Yuan, Sun, Shan, and Yang}]{tang2025unleashing}
Tang, P.; Zhang, X.; Yang, C.; Yuan, H.; Sun, J.; Shan, D.; and Yang, Z.~J. 2025.
\newblock Unleashing the Power of Visual Foundation Models for Generalizable Semantic Segmentation.
\newblock In \emph{Proceedings of the AAAI Conference on Artificial Intelligence}, volume~39, 20823--20831.

\bibitem[{Wan et~al.(2023)Wan, Huang, Lu, Gang, and Zhang}]{wan2023seaformer}
Wan, Q.; Huang, Z.; Lu, J.; Gang, Y.; and Zhang, L. 2023.
\newblock Seaformer: Squeeze-enhanced axial transformer for mobile semantic segmentation.
\newblock In \emph{The eleventh international conference on learning representations}.

\bibitem[{Wang et~al.(2019)Wang, Huang, Cheng, Zhou, Geng, and Yang}]{wang2019apolloscape}
Wang, P.; Huang, X.; Cheng, X.; Zhou, D.; Geng, Q.; and Yang, R. 2019.
\newblock The apolloscape open dataset for autonomous driving and its application.
\newblock \emph{IEEE transactions on pattern analysis and machine intelligence}.

\bibitem[{Wang et~al.(2024)Wang, Kim, Wenxuan, Xie, Ge, Chen, Li, and Luo}]{wang2024deepaccident}
Wang, T.; Kim, S.; Wenxuan, J.; Xie, E.; Ge, C.; Chen, J.; Li, Z.; and Luo, P. 2024.
\newblock Deepaccident: A motion and accident prediction benchmark for v2x autonomous driving.
\newblock In \emph{Proceedings of the AAAI Conference on Artificial Intelligence}, volume~38, 5599--5606.

\bibitem[{Wang, Liang, and Zhang(2024)}]{wang2024curriculum}
Wang, Y.; Liang, J.; and Zhang, Z. 2024.
\newblock A curriculum-style self-training approach for source-free semantic segmentation.
\newblock \emph{IEEE Transactions on Pattern Analysis and Machine Intelligence}.

\bibitem[{Xiao et~al.(2023)Xiao, Zhao, Zhang, Luo, Yu, Chen, and Yang}]{xiao2023baseg}
Xiao, X.; Zhao, Y.; Zhang, F.; Luo, B.; Yu, L.; Chen, B.; and Yang, C. 2023.
\newblock BASeg: Boundary aware semantic segmentation for autonomous driving.
\newblock \emph{Neural Networks}, 157: 460--470.

\bibitem[{Xie et~al.(2021)Xie, Wang, Yu, Anandkumar, Alvarez, and Luo}]{xie2021segformer}
Xie, E.; Wang, W.; Yu, Z.; Anandkumar, A.; Alvarez, J.~M.; and Luo, P. 2021.
\newblock SegFormer: Simple and Efficient Design for Semantic Segmentation with Transformers.
\newblock In \emph{Neural Information Processing Systems (NeurIPS)}.

\bibitem[{Yang et~al.(2024{\natexlab{a}})Yang, Gao, Qiu, Chen, Li, Dai, Chitta, Wu, Zeng, Luo et~al.}]{yang2024generalized}
Yang, J.; Gao, S.; Qiu, Y.; Chen, L.; Li, T.; Dai, B.; Chitta, K.; Wu, P.; Zeng, J.; Luo, P.; et~al. 2024{\natexlab{a}}.
\newblock Generalized predictive model for autonomous driving.
\newblock In \emph{Proceedings of the IEEE/CVF Conference on Computer Vision and Pattern Recognition}, 14662--14672.

\bibitem[{Yang et~al.(2024{\natexlab{b}})Yang, Yan, Yuan, Mi, and Tan}]{yang2024semantic}
Yang, X.; Yan, W.; Yuan, Y.; Mi, M.~B.; and Tan, R.~T. 2024{\natexlab{b}}.
\newblock Semantic segmentation in multiple adverse weather conditions with domain knowledge retention.
\newblock In \emph{Proceedings of the AAAI Conference on Artificial Intelligence}, volume~38, 6558--6566.

\bibitem[{Yang et~al.(2025)Yang, Gu, Deng, Dong, He, and Zhang}]{yang2025uawtrack}
Yang, Y.; Gu, H.; Deng, Y.; Dong, Z.; He, Z.; and Zhang, J. 2025.
\newblock UAWTrack: Universal 3D Single Object Tracking in Adverse Weather.
\newblock In \emph{Proceedings of the AAAI Conference on Artificial Intelligence}, volume~39, 9336--9344.

\bibitem[{Yang and Yang(2022)}]{yang2022deaot}
Yang, Z.; and Yang, Y. 2022.
\newblock Decoupling Features in Hierarchical Propagation for Video Object Segmentation.
\newblock In \emph{Advances in Neural Information Processing Systems (NeurIPS)}.

\bibitem[{Yu et~al.(2021)Yu, Gao, Wang, Yu, Shen, and Sang}]{yu2021bisenet}
Yu, C.; Gao, C.; Wang, J.; Yu, G.; Shen, C.; and Sang, N. 2021.
\newblock Bisenet v2: Bilateral network with guided aggregation for real-time semantic segmentation.
\newblock \emph{International Journal of Computer Vision}, 129: 3051--3068.

\bibitem[{Zhang et~al.(2022)Zhang, Huang, Luo, Chen, Wang, Liu, Yu, and Shen}]{zhang2022topformer}
Zhang, W.; Huang, Z.; Luo, G.; Chen, T.; Wang, X.; Liu, W.; Yu, G.; and Shen, C. 2022.
\newblock Topformer: Token pyramid transformer for mobile semantic segmentation.
\newblock In \emph{Proceedings of the IEEE/CVF Conference on Computer Vision and Pattern Recognition}, 12083--12093.

\bibitem[{Zhang et~al.(2023)Zhang, Carballo, Yang, and Takeda}]{zhang2023perception}
Zhang, Y.; Carballo, A.; Yang, H.; and Takeda, K. 2023.
\newblock Perception and sensing for autonomous vehicles under adverse weather conditions: A survey.
\newblock \emph{ISPRS Journal of Photogrammetry and Remote Sensing}, 196: 146--177.

\bibitem[{Zhao et~al.(2024)Zhao, Zhang, Chen, Zhao, and Tao}]{zhao2024unimix}
Zhao, H.; Zhang, J.; Chen, Z.; Zhao, S.; and Tao, D. 2024.
\newblock Unimix: Towards domain adaptive and generalizable lidar semantic segmentation in adverse weather.
\newblock In \emph{Proceedings of the IEEE/CVF Conference on Computer Vision and Pattern Recognition}, 14781--14791.

\bibitem[{Zhou et~al.(2022{\natexlab{a}})Zhou, Liu, Qiao, Xiang, and Loy}]{zhou2022domain}
Zhou, K.; Liu, Z.; Qiao, Y.; Xiang, T.; and Loy, C.~C. 2022{\natexlab{a}}.
\newblock Domain generalization: A survey.
\newblock \emph{IEEE Transactions on Pattern Analysis and Machine Intelligence}, 45(4): 4396--4415.

\bibitem[{Zhou et~al.(2022{\natexlab{b}})Zhou, Wang, Konukoglu, and Van~Gool}]{zhou2022rethinking}
Zhou, T.; Wang, W.; Konukoglu, E.; and Van~Gool, L. 2022{\natexlab{b}}.
\newblock Rethinking semantic segmentation: A prototype view.
\newblock In \emph{Proceedings of the IEEE/CVF Conference on Computer Vision and Pattern Recognition}, 2582--2593.

\end{thebibliography}
\end{document}